\pdfoutput=1 %
\documentclass{article} %
\usepackage{preprint,times}

\usepackage{amsmath,amsfonts,bm}

\def\eqref#1{equation~\ref{#1}}

\def\1{\bm{1}}

\DeclareMathAlphabet{\mathsfit}{\encodingdefault}{\sfdefault}{m}{sl}
\SetMathAlphabet{\mathsfit}{bold}{\encodingdefault}{\sfdefault}{bx}{n}

\usepackage{hyperref}
\usepackage{url}
\usepackage{booktabs}
\usepackage{multirow}
\usepackage{cleveref}
\usepackage{caption}
\usepackage{graphicx}

\usepackage{amsmath,amssymb,amsthm}
\usepackage{array}
\usepackage{tabularx}
\usepackage{algorithm}
\usepackage{algorithmic}
\usepackage{xcolor}

\theoremstyle{plain}
\newtheorem{theorem}{Theorem}[section]

\newtheorem{proposition}[theorem]{Proposition}

\definecolor{darkred}{rgb}{0.68,0.05,0.0}
\newenvironment{redlist}{%
  \begin{list}{{\color{darkred!90}\raisebox{0.15ex}{\tiny$\blacksquare$}}}{%
    \setlength{\leftmargin}{1.25em}%
    \setlength{\itemsep}{0pt}%
    \setlength{\parsep}{0pt}%
    \setlength{\topsep}{0pt}}}%
  {\end{list}}
\newcommand{\para}[1]{\vspace{0.5em}\noindent\textbf{#1}\hspace{0.1em}}

\renewcommand{\paragraph}[1]{%
  \par\vspace{0.25em}%
  \noindent\textbf{#1}
}

\headertext{}          

\title{Weight Scale Rebalancing: \\ Transferable Training Recipes for Transformers}

\author{Yikun Hou \& Naomi Graham \& Alp Yurtsever \\
Department of Mathematics and Mathematical Statistics \\
Umeå University, Sweden \\
\texttt{\{yikun.hou,naomi.graham,alp.yurtsever\}@umu.se}
}

\begin{document}

\maketitle

\begin{abstract}

Training Transformers requires careful hyperparameter tuning, which becomes increasingly costly for large models.
We introduce UDV-Rebalance (UDVR), a reparameterization that inserts an auxiliary diagonal matrix $D$ between naturally paired weight matrices $(U,V)$ appearing throughout Transformer architectures.
UDVR periodically rebalances the factors to maintain unit-norm columns in $U$ and unit-norm rows in $V$, with $D$ absorbing the corresponding scales.
With rebalancing, a single hyperparameter setting of UDVR remains effective across model scales, architectures, and data distributions.
This yields a tuning-free rule that adapts to each weight matrix using only its dimensions.
We fit this rule once on ViT models trained on CIFAR-10 and CIFAR-100, and it remains valid when the activation function or the optimizer is changed.
The same rule transfers to new image domains in DomainNet, to ViT, Swin, and MaxViT on ImageNet, and to language-model pretraining with up to 1.3B parameters.
UDVR consistently outperforms Weight Normalization and $\sigma$Reparam and matches a vanilla baseline whose hyperparameters are swept on each target model.

\end{abstract}

\section{Introduction}
\label{sec: introduction}

Training a transformer well requires choosing a learning rate and a weight decay, and the right choice shifts with every architecture and every scale \citep{yang2021tuning, zheng2026scaling}.
Each new model therefore demands its own sweep, a cost that practitioners without large compute budgets cannot afford.
Many works have studied ways to reduce this per-model sweep as much as possible. 
Hyperparameter transfer methods such as $\mu$P \citep{yang2021tuning} carry a configuration tuned on a small proxy model across width.
The proxy configuration must nevertheless be re-established for each new architecture.
The tuning is therefore relocated rather than removed.
Layer reparameterizations such as weight normalization \citep{salimans2016weight}, spectral normalization \citep{miyato2018spectral}, $\sigma$Reparam \citep{zhai2023stabilizing}, and WeSaR \citep{nishida2024initialization} stabilize training and in some cases reduce its sensitivity to hyperparameters.
However, reduced sensitivity does not tell the practitioner which hyperparameter values to use.
None of these methods provides a concrete configuration before training starts without a hyperparameter sweep on the target network.
We show that a single reparameterization, UDV-Rebalance (UDVR), yields at negligible overhead a training recipe whose hyperparameters are determined by the shape of each layer.
This recipe can therefore be fixed once and reused across architectures and scales.

Given any pair of dense layers with weights $U$ and~$V$\!, UDVR inserts an auxiliary diagonal matrix $D$ between them, yielding a $UDV$ block.
It constrains each column of $U$ and each row of $V$ to a shape-determined norm and absorbs the residual scale into $D$.
This decouples direction from magnitude and keeps the forward gain of each block stable throughout training.
Two hyperparameters are therefore introduced, the shape-determined norm scaler $\alpha$ and the corresponding learning rate constant $C$.
Their selection is extracted once, by a characterization sweep on small vision models, and this sweep is a one-off cost of establishing the rules, not part of deploying them.
We highlight that the rules are reused unchanged across matrix shapes and across task domains, so deploying UDVR on any new model requires no sweep on the real network.
UDVR thus achieves the goal of $\mu$P, a configuration that transfers across scale.
Unlike $\mu$P, it does so through a fixed shape rule rather than a proxy configuration re-tuned for each new architecture.

The rules are established on vision task, where ViT \citep{dosovitskiy2021an} families are trained from scratch on CIFAR-10 and CIFAR-100 \citep{krizhevsky2009learning}.
We then validate the rules on DomainNet \citep{peng2019moment} and ImageNet \citep{5206848}.
We further transfer them without re-tuning to language model pretraining from scratch on GPT-2-style transformers \citep{radford2019language} ranging from 0.1B to 1.3B parameters.
On CIFAR-10 and CIFAR-100, where the constants are characterized, both the swept and the predefined UDVR configurations outperform all swept baselines  on nearly all settings. 
On the sweep-free deployment targets DomainNet, ImageNet and language pretraining, it performs on par with or better than other methods in most cases.
These baselines reach their accuracy only after a hyperparameter sweep, whereas UDVR requires none.

The motivation for this approach rests on the observation that bilinear parameterizations are inherently non-identifiable; multiple choices of factor matrices can represent the same product while, at the same time, these equivalent representations can induce different optimization dynamics, including different smoothness constants and step-size requirements. This creates a dependence of optimization hyperparameters on the particular scaling of the factors, even when the represented function is unchanged. This work investigates whether explicitly controlling this scaling through rebalancing can improve hyperparameter transfer across model scales.

We summarize our contributions as follows:
\begin{redlist}
    \item \textbf{Scale-retaining reparameterization via UDV-Rebalance.}
    We insert an auxiliary diagonal scale buffer $D$ between each pair of weight matrices $(U, V)$ whose composition appears in a Transformer's forward computation, namely the query-key, value-output, and MLP projection pairs, to form $UDV$ blocks.
    The buffer is excluded from optimizer updates and changes only through prescribed rebalancing steps.
    After each optimizer update, the rebalancing step constrains the columns of $U$ and the rows of $V$ to a fixed norms and absorbs the removed scale into $D$.
    This decouples direction from magnitude while retaining the full scale of each block.
    The composed map is preserved exactly in attention blocks, and the forward gain remains stable in MLP blocks.
    Since $D$ is diagonal, each $UDV$ block adds only $O(r)$ buffer storage and element-wise scaling, where $r$ is the inner dimension, and both costs are negligible in practice.
    When every method is swept on CIFAR, UDVR outperforms the Vanilla baseline, weight normalization, and $\sigma$Reparam in nearly all settings.

    \item \textbf{A shape-determined deployment rule that transfers across Transformer architectures.}
    A one-off characterization sweep on vision Transformers trained on CIFAR establishes a rule that fixes the UDVR hyperparameters before training.
    The rule depends only on the shape of each reparameterized weight pair and therefore applies unchanged to ViT, Swin, and MaxViT at every scale, with no sweep on the target network.
    Since a decoder-only language model contains the same attention and FFN weight pairs, the rule carries over to language model pretraining without any language-specific tuning.
     At every scale, the predefined UDVR configuration outperforms the reparameterization baselines and remains competitive with the Vanilla baseline, even though all baselines are tuned on each target.
\end{redlist}

\vspace{-0.25em}
\section{Related Work}
\label{sec: related-work}
\vspace{-0.75em}

\para{Hyperparameter transfer across scale.}
$\mu$P \citep{yang2021tuning} assigns layer-wise learning rates and initialization variances so that the optimal learning rate is approximately width-invariant and a configuration tuned on a small proxy transfers to a larger model, and u-$\mu$P \citep{blake2025umup} simplifies this parameterization.
The spectral condition of \citet{yang2023spectral} gives an operator-norm account of the same effect, scaling the spectral norm of each weight matrix and each update as $\sqrt{d_{\text{out}}/d_{\text{in}}}$ and thereby recovering $\mu$P.
A recent survey \citep{pethick2025training} unifies these views through the operator norm and notes that, although layer shape fixes how each layer's norm constraint scales with width, the induced Lipschitz constants are too loose to set the hyperparameters directly, which are instead chosen by search.
Our deployment rule shares this operator-norm viewpoint but obtains a width-invariant per-layer update scale from a measured shape rule rather than an analytical derivation.

\para{Magnitude-direction decoupling.}
Separating a weight's magnitude from its direction is known to be an effective strategy to reduce hyperparameter sensitivity and has been approached in a few different ways.
Weight normalization \citep{salimans2016weight} splits a weight vector into an independently optimized direction and norm.
Alternatively, one can fix the norm and optimize direction alone, as in nGPT \citep{loshchilov2025ngpt}, which constrains weights to the unit sphere.
MD Decoupling \citep{hagele2026improving} treats magnitude as a separate variable with its own dynamics, letting the two carry different learning rates, and \citet{velikanov2026learnable} instead treat magnitude as an additional scale to be learned through explicit multipliers.
Each of these approaches reduces hyperparameter sensitivity, yet none fixes the learning rate in advance.
The learning rate is still found by search, either directly on the target network or on a small proxy whose optimum then transfers across width.

\para{Norm and spectral constraints.}
A separate family of methods control magnitude through a norm or spectral constraint to stabilize training.
Spectral normalization \citep{miyato2018spectral} bounds the largest singular value for Lipschitz control, $\sigma$Reparam \citep{zhai2023stabilizing} pairs spectral normalization with a learned scalar to prevent attention entropy collapse, and WeSaR \citep{nishida2024initialization} equalizes weight norms through a per-matrix gate to mitigate loss spikes in language model pretraining.
Each of these constraints stabilizes training, yet none of them prescribes the hyperparameters, which are still left to the practitioner.
UDVR instead accumulates the magnitude in $D$ through rebalancing, which requires no learning rate of its own, and prescribes the learning rate of $U$ and $V$ through the shape rule.

\para{Geometric reparameterizations of bilinear models.}
Recent works have explored the direction-magnitude decompositions for low-rank matrix optimization. 
PoLAR \citep{lion2026polar} represents low-rank updates using two orthonormal direction matrices and an intermediate magnitude matrix, employing Riemannian optimization to improve the utilization of the available rank. 
Similarly, Direction–Magnitude Decomposition (DMD) \citep{wei2026direction} separates directional and magnitude components to improve the convergence of low-rank matrix optimization. 
Stiefel-AdamW \citep{zangrando2026stiefel} takes a complementary approach by constraining one factor in a bilinear neural-network layer to the Stiefel manifold and incorporating this geometry into AdamW. 
These approaches target the gauge symmetry present in linear factorization blocks $W=UV$ by replacing the original unconstrained factorization with parameterizations involving orthonormal factors. 
In contrast, our approach retains the original factorization and exploits its scaling symmetry without imposing orthonormality constraints. 
While these works address several aspects of low-rank optimization, none explicitly addresses hyperparameter transfer across model sizes, which is the focus of our work.

\section{Methodology}
\label{sec: methodology}

Transformer blocks \citep{vaswani2017attention} contain natural $(U,V)$ factor pairs.
Following \citet{elhage2021mathematical}, each attention head yields the pairs $(W_q^i,{W_k^i}^\top)$ and $(W_v^i,W_o^i)$, and the FFN yields $(W_{\text{up}},W_{\text{down}})$ (see \Cref{app:udv_pairs} for details).
Each pair has $U \in \mathbb{R}^{d \times r}$ and $V \in \mathbb{R}^{r \times d}$, where $d$ is the embedding dimension and $r$ is $d_{\text{head}}$ for attention and $d_{\text{ffn}}$ for the FFN.
A diagonal matrix $D \in \mathbb{R}^{r \times r}$ is inserted between $U$ and $V$.
Importantly, $D$ evolves only through the rebalance rule defined below and is excluded from gradient-based optimizer updates.

\para{UDV-Rebalance (UDVR).}
We separate the directions and magnitudes of the weight factors by maintaining a fixed Euclidean norm for the columns of $U$ and the rows of $V$, while capturing their scales by the diagonal matrix $D$.
After each gradient step, we enforce these constraints through a \textit{rebalance} operation: every column of $U$ and every row of $V$ is rescaled to the fixed norm $\alpha\sqrt{d}$, and the stripped scale is moved into $D$.
For $X \in \mathbb{R}^{d \times r}$, let $\mathcal{S}(X)$ be the diagonal matrix with $\mathcal{S}(X)[k,k] = \alpha\sqrt{d}\,/\,\lVert X[:,k] \rVert_2$.
With $\mathcal{L}$ the training objective and $\eta_t$ the learning rate of the pair at step $t$, one step reads
\begin{equation}
\begin{aligned}
\tilde U_{t+1} & = U_t - \eta_t \nabla_U \mathcal{L}, \\
\tilde V_{t+1} & = V_t - \eta_t \nabla_V \mathcal{L}, \\
U_{t+1} & = \tilde U_{t+1} S^U_{t+1}, \\
V_{t+1} & = S^V_{t+1} \tilde V_{t+1}, \\
D_{t+1} & = (S^U_{t+1})^{-1} D_t\, (S^V_{t+1})^{-1},
\end{aligned}
\label{eq:rebalance}
\end{equation}
where $S^U_{t+1} = \mathcal{S}(\tilde U_{t+1})$ and $S^V_{t+1} = \mathcal{S}(\tilde V_{t+1}^{\top})$.
Since $D_{t+1}$ absorbs exactly the rescaling,
we have $U_{t+1} D_{t+1} V_{t+1} = \tilde U_{t+1} D_t \tilde V_{t+1}$.
Hence the rebalance fixes
$\lVert U \rVert_F = \lVert V \rVert_F = \alpha\sqrt{dr}$.

For the attention pairs, $U$ and $V$ interact only through a linear operation, so their effect on the network is fully captured by the composed map $UDV$, which the rebalance leaves exactly invariant.
For the FFN, the rebalance is exactly function-preserving when the activation is positively homogeneous (e.g., ReLU), since the positive diagonal rescalings it produces commute with the activation.
For non-homogeneous activations such as GELU this no longer holds, we therefore drop the contribution of $V$ to the update of $D$.
In the FFN, $D_{t+1} = (S^U_{t+1})^{-1} D_t$, and $V$ is only projected back onto its sphere after every step.
Pseudocode for UDVRebalance is provided in \Cref{app:pseudocode}.
To demonstrate that the rebalancing step does not interfere with the usual convergence rates, \Cref{app:proofs} establishes that, in the idealized case where $F(U,D,V)=f(UDV)$, using stochastic gradient descent with a constant step size $\eta=\Theta(T^{-1/2})$ results in the usual $O(T^{-1/2})$ averaged stationarity rate.

Without an additional diagonal matrix $D$ in between a product $UV$, setting both $U$ and $V$ to have a fixed norm requires estimating the appropriate norm of the optimal (or near-optimal) product $UV$. This norm should scale with matrix dimension and model depth. However, we do not have access to a predefined rule to determine the scale a priori.
To avoid the challenge and possible error of estimating it, we allow the matrix $D$ to vary without any strict bounds, so that the norm of product $UDV$ can grow into whatever scale the optimization objective requires. 
By fixing the norms of the columns of $U$ and the rows of $V$ throughout optimization, we effectively treat them as the incoming and outgoing directions of individual intermediate neurons. The corresponding magnitudes are then carried by $D$, which is neither fixed nor directly updated by the optimizer, but evolves dynamically through rebalancing during training.

\para{Regularizer.}
Nothing in the rebalance bounds $D$, and without a constraint the forward operator norm of a block grows without bound and training fails.
We therefore penalize $D$ with the same strength in attention and in the FFN.
$D$ receives no gradient, so the penalty acts through the factors instead. 
Its gradient scales each factor down without changing its direction, and the next rebalance moves that shrinkage into $D$.
In attention both factors pass their shrinkage to $D$, whereas in the FFN only $U$ does, since the stripped scale of $V$ is discarded.
We therefore impose the penalty as
\begin{equation}
\begin{aligned}
\mathcal{R}_{\text{attn}} &= \gamma\big(\lVert UD\rVert_F^2 + \lVert DV\rVert_F^2\big),\\
\mathcal{R}_{\text{ffn}}  &= 2\gamma\big(\lVert UD\rVert_F^2 + \lVert V\rVert_F^2\big).
\end{aligned}
\label{eq:regularizer}
\end{equation}
Attention splits the penalty between $U$ and $V$, while the FFN places all of it on $U$.
The $V$ term in $\mathcal{R}_{\text{ffn}}$ is constant on the sphere and is kept only so that $\mathcal{R}_{\text{ffn}}$ is proportional to the sum of squared norms of the effective FFN weights $UD$ and $V$.
Since every column of $U$ and every row of $V$ is constrained to norm $\alpha\sqrt{d}$, both regularizers reduce to $2\gamma\alpha^{2}d\,\lVert D\rVert_F^2$ up to an additive constant, which determines the factor of $2$ in $\mathcal{R}_{\text{ffn}}$ without tuning.
Writing the FFN term in the attention form would match $\mathcal{R}_{\text{ffn}}$ on the sphere up to a constant, but the rebalance would discard its $V$ half and leave $D$ under half the intended penalty (\Cref{app:pseudocode}).

\para{Learning rates.}
Parameters outside UDV blocks, such as embeddings, normalization layers, and the output head, train at a single rate $\eta_{\text{main}}$.
The factors $U$ and $V$ take a rate depends on their shape.
We scale the rate of each factor $W \in \{ U, V\}$ with the fixed norm imposed by the rebalance, normalized by the width so that it does not grow as the model widens,
\begin{equation}
\eta_W = C\,\lVert W \rVert_F \,/\, d,
\label{eq:lr-rule}
\end{equation}
with one constant $C$ for all factors.
Substituting $\lVert W \rVert_F = \alpha\sqrt{dr}$ from the rebalance, with $r = h\,d_{\text{head}} = d$ for an attention factor stacked over its $h$ heads and $r = d_{\text{ffn}}$ for an FFN factor, gives
\begin{equation}
\eta_{\text{attn}} = C\alpha,
\qquad
\eta_{\text{ffn}} = \sqrt{d_{\text{ffn}}/d}\;C\alpha .
\label{eq:width-invariant}
\end{equation}
Both are independent of $d$, and their ratio is set by the architecture alone (e.g., $\eta_{\text{ffn}} = 2\,\eta_{\text{attn}}$ for the standard $d_{\text{ffn}} = 4d$).

\section{Experiments}
\label{sec: experiments}

We evaluate our method on image classification and extend to the language-model pretraining, both with Transformer-based models.
All methods share an identical evaluation protocol, including architecture, data, training budget, augmentation, and random seeds. 
Full experimental details are provided in \Cref{app:exp_details}.

To assess the proposed method, we compare UDVR against three baselines, a vanilla model with no reparameterization, weight normalization \citep{salimans2016weight}, and $\sigma$Reparam \citep{zhai2023stabilizing}, on image classification and language modeling.
Each task uses a family of model scales, listed with their trainable parameter counts in \Cref{tab:models}.
In vision, we use ViT \citep{dosovitskiy2021an} with configurations following DeiT \citep{touvron2021training}, together with Swin \citep{liu2021swin} and MaxViT \citep{tu2022maxvit}. 
For language modeling, we adopt a GPT-2-style architecture whose scales follow the Pythia family \citep{biderman2023pythia}.

All baselines are tuned separately for each model.
The vanilla model is tuned over a grid of learning rates and weight decays.
Weight normalization and $\sigma$Reparam are tuned over the learning rate only.
Following their original recipes, both methods are trained without weight decay as it would alter their reparameterization dynamics.
\Cref{tab:cv_sweep_common} summarizes the CIFAR sweep ranges for these baselines.
In contrast, every UDVR result in the main paper uses one fixed configuration.
The constrained scale $\alpha = 10^{-2}$ and the regularizer strength $\gamma = 10^{-7}$ are fixed after a sweep in (\Cref{app:hparams}).
The only hyperparameter we select is the learning-rate constant $C$.
\emph{We determine $C = 0.0217$ once on CIFAR-10 and CIFAR-100 with ViT and reuse it unchanged for every architecture and dataset}.
We obtain this value by a grid search over $C \in \{0.0109, 0.0154, 0.0217, 0.0307, 0.0434\}$ and keep the value whose accuracy varies least across ViT models of different sizes.
The non-UDV parameters use a fixed learning rate without per-model tuning, and no parameter receives weight decay.
The selection of all hyperparameters is detailed in \Cref{app:hparams}.

\begin{table}[t]
\centering
\small
\setlength{\tabcolsep}{3pt}
\caption{Model scales and trainable parameter counts. For Swin and MaxViT, $d$ gives the widths from stage 1 to stage 4. All models use $d_{\text{ffn}}=4d$ and $d_{\text{head}}=64$ (32 for Swin and MaxViT).}
\begin{tabular}{llccrrr}
\toprule
Domain & Data & Model & $d$ & \multicolumn{3}{c}{Params (M)} \\
\midrule
\multirow{11}{*}{\shortstack[l]{Vision}}
 & \multirow{5}{*}{\shortstack[l]{CIFAR\\DomainNet}}
   & & & $L{=}6$ & $L{=}8$ & $L{=}12$ \\
\cmidrule(lr){5-7}
 & & ViT & 128 & 1.21 & 1.60 & 2.39 \\
 & & ViT & 192 & 2.70 & 3.59 & 5.36 \\
 & & ViT & 256 & 4.78 & 6.35 & 9.50 \\
\cmidrule(lr){2-7}
 & \multirow{6}{*}{ImageNet}
   & ViT-S    & 384       & \multicolumn{3}{c}{22.01} \\
 & & ViT-B    & 768       & \multicolumn{3}{c}{86.48} \\
 & & Swin-T   & 96--768   & \multicolumn{3}{c}{28.25} \\
 & & Swin-B   & 128--1024 & \multicolumn{3}{c}{87.66} \\
 & & MaxViT-T & 64--512   & \multicolumn{3}{c}{30.60} \\
 & & MaxViT-S & 96--768   & \multicolumn{3}{c}{68.26} \\
\midrule
\multirow{3}{*}{Language}
 & \multirow{3}{*}{\shortstack[l]{SlimPajama}}
   & GPT-S & 512  & \multicolumn{3}{c}{120.76} \\
 & & GPT-M   & 1024 & \multicolumn{3}{c}{379.94} \\
 & & GPT-L & 2048 & \multicolumn{3}{c}{1313.53} \\
\bottomrule
\end{tabular}
\label{tab:models}
\end{table}

\begin{table}[htb!]
\centering
\small
\caption{CIFAR sweep grids, given as endpoints and number of grid points. A single value denotes a fixed setting. UDVR uses a single configuration, and its learning rate refers to the non-UDV parameters. On ImageNet, all learning rates except those of the UDV parameters are scaled linearly with the batch size \citep{goyal2017accurate,touvron2021training}.}
\begin{tabular}{lcc}
\toprule
Method & Learning rate & Weight decay\\
\midrule
Vanilla         & $[10^{-4}, 10^{-2}]$ (5) & $[0, 10^{-1}]$ (6) \\
WN              & $[10^{-5}, 10^{-1}]$ (9) & $0$ \\
$\sigma$Reparam & $[10^{-5}, 10^{-1}]$ (9) & $0$ \\
UDVR            & $3{\times}10^{-4}$       & $0$ \\
\bottomrule
\end{tabular}
\label{tab:cv_sweep_common}
\end{table}

\subsection{Vision task}
\label{sec:cv}

\para{Setup.}
We train the vision models listed in \Cref{tab:models}~from scratch on CIFAR-10 (200 epochs) and CIFAR-100 (300 epochs) \citep{krizhevsky2009learning}, and on ImageNet-1K (300 epochs) \citep{5206848}.
All methods use AdamW \citep{loshchilov2018decoupled} with a cosine learning-rate schedule that decays to zero after a $5\%$ linear warmup.
The batch size is 512 on CIFAR and 2048 (global) on ImageNet.
Following DeiT \citep{touvron2021training} and common practice for ViT training, we apply AutoAugment \citep{cubuk2018autoaugment}, mixup \citep{zhang2018mixup}, CutMix \citep{yun2019cutmix}, label smoothing \citep{szegedy2016rethinking}, and stochastic depth \citep{huang2016deep}.
We report top-1 test accuracy, together with top-5 accuracy on ImageNet, averaged over the last 5 epochs, with 3 seeds on CIFAR and a single seed on ImageNet.
The configuration of UDVR is selected on CIFAR and kept fixed in all subsequent experiments.
On CIFAR, we additionally examine its robustness to the optimizer, using Lion \citep{chen2023symbolic} and Muon \citep{jordan2024muon}, and to the activation function, using ReLU \citep{nair2010rectified} and SiLU \citep{elfwing2018sigmoid}.
We then evaluate it under data distribution shift on DomainNet \citep{peng2019moment}.
Finally, we extend it to ImageNet with larger ViT models as well as Swin \citep{liu2021swin} and MaxViT \citep{tu2022maxvit}.
On ImageNet, the baseline sweeps on the larger models are narrower and centered on the optimum found for the smaller model of the same architecture.

\para{CIFAR results.}
\Cref{tab:cifar_tuned_baselines} compares UDVR against Vanilla, WN, and $\sigma$Reparam, each swept to its own optimum for every model on CIFAR-10 and CIFAR-100.
In contrast, UDVR uses a single fixed configuration derived from a hyperparameter sweep for all models.
For a few models, the per-model optimal $C$ differs from the shared value.
The reported UDVR accuracy is therefore conservative relative to per-model tuning (\Cref{app:hparams}).

\begin{table}[t]
\centering
\small
\caption{Test accuracy on CIFAR-10 and CIFAR-100.}
\begin{tabular}{cc cccc cccc}
\toprule
 & & \multicolumn{4}{c}{CIFAR-10} & \multicolumn{4}{c}{CIFAR-100} \\
\cmidrule(lr){3-6}\cmidrule(lr){7-10}
$d$ & $L$ & UDVR & Vanilla & WN & $\sigma$Reparam & UDVR & Vanilla & WN & $\sigma$Reparam \\
\midrule
\multirow{3}{*}{128} & 6  & \textbf{87.96} & 85.15 & 83.05 & 83.55 & \textbf{65.39} & 64.55 & 62.14 & 62.62 \\
                     & 8  & \textbf{88.54} & 86.05 & 84.33 & 83.32 & \textbf{66.62} & 65.17 & 62.32 & 62.77 \\
                     & 12 & \textbf{88.62} & 86.96 & 85.04 & 84.37 & \textbf{68.16} & 66.78 & 64.18 & 64.83 \\
\midrule
\multirow{3}{*}{192} & 6  & \textbf{87.72} & 86.63 & 85.16 & 84.20 & \textbf{67.19} & 66.47 & 63.38 & 64.03 \\
                     & 8  & \textbf{88.43} & 87.11 & 85.54 & 85.12 & \textbf{68.17} & 65.92 & 64.28 & 64.66 \\
                     & 12 & \textbf{88.67} & 87.12 & 86.07 & 85.76 & \textbf{69.07} & 67.85 & 66.76 & 66.67 \\
\midrule
\multirow{3}{*}{256} & 6  & \textbf{88.19} & 87.61 & 86.54 & 85.67 & \textbf{68.36} & 67.28 & 66.07 & 65.37 \\
                     & 8  & \textbf{89.15} & 87.61 & 86.57 & 85.73 & \textbf{69.83} & 67.62 & 66.60 & 66.92 \\
                     & 12 & \textbf{89.70} & 88.76 & 87.19 & 86.92 & \textbf{68.79} & 67.91 & 67.74 & 67.09 \\
\midrule
\multicolumn{2}{c}{Mean} & \textbf{88.55} & 87.00 & 85.50 & 84.96 & \textbf{67.95} & 66.62 & 64.83 & 65.00 \\
\bottomrule
\end{tabular}%
\label{tab:cifar_tuned_baselines}
\end{table}

\begin{figure*}[htb!]
    \centering
    \includegraphics[width=\textwidth]{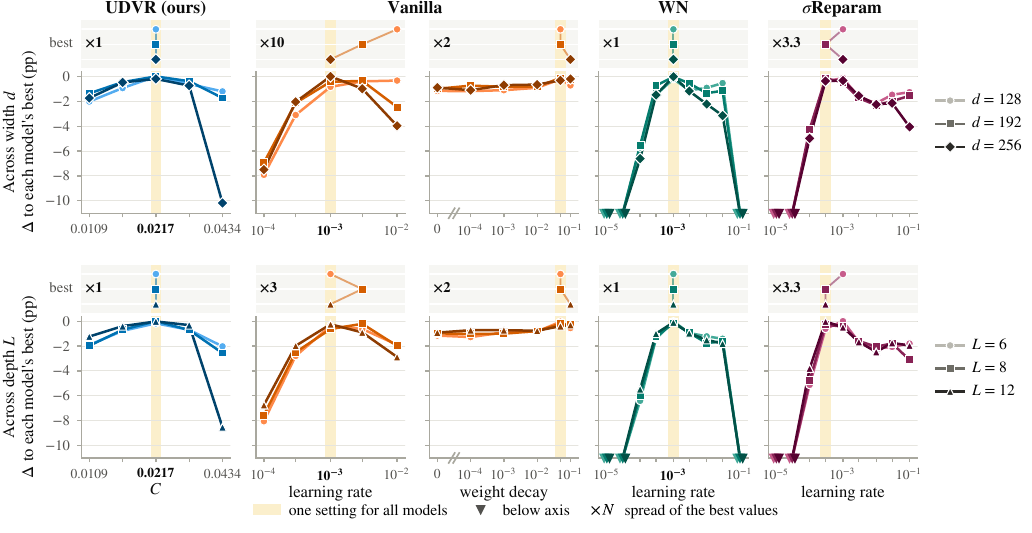}
    \caption{\textbf{Shift of the optimal hyperparameter with model width and depth on CIFAR.}
    Each column shows one hyperparameter of one method.
    For Vanilla, each learning rate is paired with the best weight decay of each run, and each weight decay with the best learning rate.
    In the top row, each point averages the accuracy drop from the best setting over the three depths and two CIFAR datasets of one width.
    The bottom row does the same over the three widths and two datasets of one depth.
    A line therefore reaches zero only when all its runs share the same optimum.
    The strip above each panel marks the hyperparameter value at which each line peaks.
    $\times N$ is the ratio between the largest and smallest of these peak values within the panel.
    Shading marks the single value with the smallest average drop across all runs.}
    \label{fig:cifar_hyper_robust}
\end{figure*}

\Cref{fig:cifar_hyper_robust} tracks how the optimal hyperparameter of each method shifts as the model grows.
For Vanilla, the optimal learning rate shifts with width ($\times 10$), even though the weight decay is retuned for every model.
This trend implies that the learning rate has to be searched again whenever the width changes.
The optimal weight decay, in contrast, only alternates between two adjacent grid values without a trend ($\times 2$).
For UDVR, all widths and depths share the same optimal $C^\star$ ($\times 1$ along both axes).
A constant selected once therefore transfers to new model shapes without re-tuning.
Such a stable optimum is also observed for WN, which keeps a single optimal learning rate ($\times 1$), and for $\sigma$Reparam, whose optimum moves by one grid step without a monotone trend across widths ($\times 3.3$).
UDVR therefore does not stand out from these two reparameterization baselines through a more consistent optimal $C$ across models.
Its advantage lies in the higher accuracy it reaches at $C^\star$ (\Cref{tab:cifar_tuned_baselines}).
The observations along the depth axis are consistent with those along the width axis, except that the optimal learning rate of Vanilla shows no monotone trend with depth ($\times 3$).

To evaluate whether the rule is robust, we change one component of the training setup at a time and keep the rest unchanged.
\begin{redlist}
    \item \textbf{Optimizers.}
    We examine the effect of the optimizer by replacing AdamW with Lion and Muon.
    For every method, the learning rate is converted from its AdamW setting rather than searched again.
    We match the per-element root-mean-square (RMS) of each Lion or Muon update to $0.2\eta$, the typical RMS of an AdamW update with learning rate $\eta$ \citep{liu2025muon}.
    This gives the learning rate $0.2\eta$ for Lion, whose sign update has unit RMS \citep{chen2023symbolic}, and $0.2\eta\sqrt{\max(m,n)}$ for Muon on an $m \times n$ weight matrix.
    On CIFAR-100, UDVR achieves the highest mean accuracy under both optimizers and the best accuracy on 8 of 9 models under each of them (\Cref{tab:cifar100_optimizers}).
    The only exception is the narrowest and deepest model with $d{=}128$ and $L{=}12$, where Vanilla is more accurate under both optimizers.
    On CIFAR-10, UDVR also attains the highest mean accuracy under both optimizers (\Cref{tab:cifar10_optimizers}).
    
    \item \textbf{Activations.}
    The FFN in ViT uses GELU \citep{hendrycks2016gaussian} by default, and we replace it with ReLU and SiLU while all methods reuse their GELU hyperparameters.
    On CIFAR-100, UDVR achieves the best accuracy on all 18 models (\Cref{tab:cifar100_activations}).
    Switching from ReLU to SiLU lowers the mean accuracy of every baseline by 1.6 to 2.5 points, while that of UDVR remains essentially unchanged.
    On CIFAR-10, UDVR is the best on 17 of 18 models and is equally stable under SiLU (\Cref{tab:cifar10_activations}).
    
    \item \textbf{Datasets.} We next test whether the setting chosen on CIFAR survives a change of data distribution.
    Each network is trained on the Real and Sketch domains of DomainNet without re-tuning any method.
    Real contains natural photographs like CIFAR, whereas Sketch renders its categories as line drawings and thus represents a larger shift.
    The baselines reuse the hyperparameters selected for each structure on CIFAR-100, while UDVR keeps the fixed $C$ for all structures.
    Under this protocol, UDVR achieves the highest accuracy at every width and depth on both domains (\Cref{tab:domainnet}).
    The baselines transfer less reliably, and WN even diverges on Sketch at $d{=}128$ and $L{=}8$ with the learning rate that works on Real.
    The margin of UDVR over the strongest baseline grows from 2.8 points on Real to 8.5 points on Sketch, indicating that the tuning-free rule calibrated on CIFAR remains robust and gains the most under the larger domain shift.
    
\end{redlist}

\begin{table}[t]
\centering
\small
\caption{Test accuracy on CIFAR-100 under Lion and Muon.}
\resizebox{\columnwidth}{!}{%
\begin{tabular}{cc cccc cccc}
\toprule
 & & \multicolumn{4}{c}{Lion} & \multicolumn{4}{c}{Muon} \\
\cmidrule(lr){3-6}\cmidrule(lr){7-10}
$d$ & $L$ & UDVR & Vanilla & WN & $\sigma$Reparam & UDVR & Vanilla & WN & $\sigma$Reparam \\
\midrule
\multirow{3}{*}{128} & 6  & \textbf{67.54} & 65.07 & 64.25 & 63.05 & \textbf{69.10} & 68.44 & 65.09 & 65.81 \\
                     & 8  & \textbf{67.43} & 66.11 & 63.05 & 64.25 & \textbf{70.29} & 70.26 & 65.56 & 67.90 \\
                     & 12 & 68.55 & \textbf{69.32} & 66.23 & 65.17 & 71.43 & \textbf{74.08} & 67.52 & 67.32 \\
\midrule
\multirow{3}{*}{192} & 6  & \textbf{68.82} & 67.25 & 65.57 & 65.31 & \textbf{70.09} & 69.92 & 67.20 & 67.16 \\
                     & 8  & \textbf{69.13} & 67.88 & 65.22 & 65.69 & \textbf{71.02} & 70.77 & 67.38 & 66.68 \\
                     & 12 & \textbf{69.94} & 68.79 & 67.49 & 67.01 & \textbf{71.78} & 70.44 & 68.92 & 67.90 \\
\midrule
\multirow{3}{*}{256} & 6  & \textbf{68.71} & 67.52 & 65.96 & 66.15 & \textbf{69.60} & 68.41 & 67.19 & 67.78 \\
                     & 8  & \textbf{69.91} & 67.78 & 66.84 & 66.79 & \textbf{70.53} & 69.55 & 68.22 & 66.22 \\
                     & 12 & \textbf{70.32} & 69.35 & 68.00 & 67.79 & \textbf{71.47} & 70.66 & 69.58 & 70.53 \\
\midrule
\multicolumn{2}{c}{Mean} & \textbf{68.93} & 67.67 & 65.85 & 65.69 & \textbf{70.59} & 70.28 & 67.41 & 67.48 \\
\bottomrule
\end{tabular}%
}
\label{tab:cifar100_optimizers}
\end{table}

\begin{table}[t]
\centering
\small
\caption{Test accuracy on CIFAR-10 under Lion and Muon.}
\resizebox{\columnwidth}{!}{%
\begin{tabular}{cc cccc cccc}
\toprule
 & & \multicolumn{4}{c}{Lion} & \multicolumn{4}{c}{Muon} \\
\cmidrule(lr){3-6}\cmidrule(lr){7-10}
$d$ & $L$ & UDVR & Vanilla & WN & $\sigma$Reparam & UDVR & Vanilla & WN & $\sigma$Reparam \\
\midrule
\multirow{3}{*}{128} & 6  & \textbf{88.13} & 66.94 & 84.95 & 82.65 & \textbf{90.66} & 90.59 & 87.79 & 88.19 \\
                     & 8  & \textbf{88.04} & 87.01 & 85.39 & 83.60 & 90.64 & \textbf{91.03} & 88.51 & 89.33 \\
                     & 12 & \textbf{88.61} & 87.34 & 85.90 & 84.98 & 91.35 & \textbf{91.87} & 88.81 & 88.51 \\
\midrule
\multirow{3}{*}{192} & 6  & \textbf{87.77} & 87.26 & 86.09 & 83.12 & \textbf{91.00} & 90.18 & 89.61 & 89.67 \\
                     & 8  & \textbf{88.59} & 85.34 & 86.78 & 85.32 & 91.94 & \textbf{91.96} & 90.31 & 88.34 \\
                     & 12 & \textbf{88.72} & 87.73 & 86.95 & 86.65 & \textbf{92.10} & 91.30 & 90.94 & 89.49 \\
\midrule
\multirow{3}{*}{256} & 6  & \textbf{88.04} & 88.02 & 86.21 & 85.78 & \textbf{91.15} & 90.35 & 90.16 & 89.86 \\
                     & 8  & 88.39 & \textbf{88.53} & 87.88 & 85.91 & \textbf{91.89} & 91.05 & 90.81 & 90.72 \\
                     & 12 & \textbf{89.20} & 88.88 & 87.84 & 87.31 & 92.17 & \textbf{92.41} & 91.37 & 89.79 \\
\midrule
\multicolumn{2}{c}{Mean} & \textbf{88.39} & 85.23 & 86.44 & 85.04 & \textbf{91.43} & 91.19 & 89.81 & 89.32 \\
\bottomrule
\end{tabular}%
}
\label{tab:cifar10_optimizers}
\end{table}

\begin{table}[t]
\centering
\small
\caption{Test accuracy on CIFAR-100 under ReLU and SiLU.}
\resizebox{\columnwidth}{!}{%
\begin{tabular}{cc cccc cccc}
\toprule
 & & \multicolumn{4}{c}{ReLU} & \multicolumn{4}{c}{SiLU} \\
\cmidrule(lr){3-6}\cmidrule(lr){7-10}
$d$ & $L$ & UDVR & Vanilla & WN & $\sigma$Reparam & UDVR & Vanilla & WN & $\sigma$Reparam \\
\midrule
\multirow{3}{*}{128} & 6  & \textbf{65.80} & 63.56 & 62.47 & 62.34 & \textbf{65.59} & 62.91 & 58.79 & 59.59 \\
                     & 8  & \textbf{66.60} & 64.06 & 60.83 & 63.19 & \textbf{66.27} & 63.83 & 60.34 & 60.32 \\
                     & 12 & \textbf{67.90} & 65.27 & 64.64 & 64.42 & \textbf{67.83} & 63.27 & 60.55 & 62.62 \\
\midrule
\multirow{3}{*}{192} & 6  & \textbf{66.92} & 64.87 & 64.05 & 63.93 & \textbf{67.21} & 63.72 & 61.31 & 62.37 \\
                     & 8  & \textbf{67.45} & 65.93 & 64.40 & 64.20 & \textbf{67.98} & 63.95 & 61.67 & 62.52 \\
                     & 12 & \textbf{69.05} & 67.03 & 65.71 & 65.24 & \textbf{69.34} & 65.61 & 63.94 & 64.58 \\
\midrule
\multirow{3}{*}{256} & 6  & \textbf{67.88} & 66.81 & 65.45 & 64.05 & \textbf{67.50} & 64.38 & 62.87 & 62.98 \\
                     & 8  & \textbf{68.66} & 67.21 & 65.96 & 65.20 & \textbf{68.60} & 65.23 & 63.99 & 64.51 \\
                     & 12 & \textbf{69.79} & 68.47 & 67.19 & 66.53 & \textbf{70.11} & 65.92 & 65.01 & 65.30 \\
\midrule
\multicolumn{2}{c}{Mean} & \textbf{67.78} & 65.91 & 64.52 & 64.34 & \textbf{67.83} & 64.31 & 62.05 & 62.75 \\
\bottomrule
\end{tabular}%
}
\label{tab:cifar100_activations}
\end{table}

\begin{table}[t]
\centering
\small
\caption{Test accuracy on CIFAR-10 under ReLU and SiLU.}
\resizebox{\columnwidth}{!}{%
\begin{tabular}{cc cccc cccc}
\toprule
 & & \multicolumn{4}{c}{ReLU} & \multicolumn{4}{c}{SiLU} \\
\cmidrule(lr){3-6}\cmidrule(lr){7-10}
$d$ & $L$ & UDVR & Vanilla & WN & $\sigma$Reparam & UDVR & Vanilla & WN & $\sigma$Reparam \\
\midrule
\multirow{3}{*}{128} & 6  & \textbf{87.45} & 85.20 & 85.10 & 84.44 & \textbf{86.88} & 83.43 & 79.64 & 78.13 \\
                     & 8  & \textbf{87.60} & 85.81 & 85.22 & 84.93 & \textbf{87.76} & 83.19 & 79.82 & 78.64 \\
                     & 12 & \textbf{88.47} & 87.27 & 84.36 & 85.55 & \textbf{88.17} & 83.73 & 83.99 & 80.38 \\
\midrule
\multirow{3}{*}{192} & 6  & \textbf{87.78} & 87.12 & 86.17 & 85.65 & \textbf{88.34} & 83.62 & 82.70 & 80.22 \\
                     & 8  & \textbf{88.20} & 87.10 & 87.29 & 86.13 & \textbf{88.21} & 84.88 & 82.91 & 81.38 \\
                     & 12 & \textbf{88.44} & 87.80 & 86.22 & 87.19 & \textbf{89.44} & 85.20 & 83.87 & 83.59 \\
\midrule
\multirow{3}{*}{256} & 6  & \textbf{88.19} & 87.82 & 86.08 & 86.91 & \textbf{89.11} & 85.11 & 84.54 & 82.71 \\
                     & 8  & 88.06 & \textbf{88.15} & 87.75 & 87.04 & \textbf{89.46} & 85.72 & 84.73 & 83.07 \\
                     & 12 & \textbf{88.78} & 88.48 & 87.71 & 87.46 & \textbf{89.82} & 85.12 & 84.76 & 84.60 \\
\midrule
\multicolumn{2}{c}{Mean} & \textbf{88.11} & 87.19 & 86.21 & 86.14 & \textbf{88.58} & 84.44 & 83.00 & 81.41 \\
\bottomrule
\end{tabular}%
}
\label{tab:cifar10_activations}
\end{table}

\begin{table}[t]
\centering
\small
\caption{Test accuracy on DomainNet Real and Sketch. $\times$ marks a run that diverged during training.}
\resizebox{\columnwidth}{!}{%
\begin{tabular}{cc cccc cccc}
\toprule
 & & \multicolumn{4}{c}{Real} & \multicolumn{4}{c}{Sketch} \\
\cmidrule(lr){3-6}\cmidrule(lr){7-10}
$d$ & $L$ & UDVR & Vanilla & WN & $\sigma$Reparam & UDVR & Vanilla & WN & $\sigma$Reparam \\
\midrule
\multirow{3}{*}{128} & 6  & \textbf{55.50} & 54.12 & 49.26 & 50.35 & \textbf{34.08} & 27.22 & 21.67 & 20.49 \\
                     & 8  & \textbf{56.69} & 54.72 & 49.93 & 50.84 & \textbf{35.68} & 29.26 & $\times$ & 20.81 \\
                     & 12 & \textbf{58.46} & 55.31 & 51.74 & 53.84 & \textbf{36.29} & 25.91 & 23.44 & 25.92 \\
\midrule
\multirow{3}{*}{192} & 6  & \textbf{57.90} & 55.63 & 52.34 & 52.48 & \textbf{37.00} & 26.77 & 23.80 & 21.76 \\
                     & 8  & \textbf{58.87} & 55.98 & 51.82 & 54.30 & \textbf{37.65} & 28.97 & 24.23 & 28.14 \\
                     & 12 & \textbf{60.45} & 55.98 & 54.11 & 55.80 & \textbf{38.48} & 27.40 & 26.60 & 30.65 \\
\midrule
\multirow{3}{*}{256} & 6  & \textbf{58.28} & 55.49 & 53.68 & 53.53 & \textbf{37.02} & 29.47 & 26.53 & 25.00 \\
                     & 8  & \textbf{59.72} & 56.80 & 54.30 & 55.00 & \textbf{38.38} & 31.74 & 29.61 & 30.77 \\
                     & 12 & \textbf{61.14} & 58.20 & 56.68 & 56.01 & \textbf{39.43} & 30.75 & 30.19 & 28.03 \\
\midrule
\multicolumn{2}{c}{Mean} & \textbf{58.56} & 55.80 & 52.65 & 53.57 & \textbf{37.11} & 28.61 & 25.76 & 25.73 \\
\bottomrule
\end{tabular}%
}
\label{tab:domainnet}
\end{table}

\begin{figure*}[htb!]
  \centering
  \includegraphics[width=\textwidth]{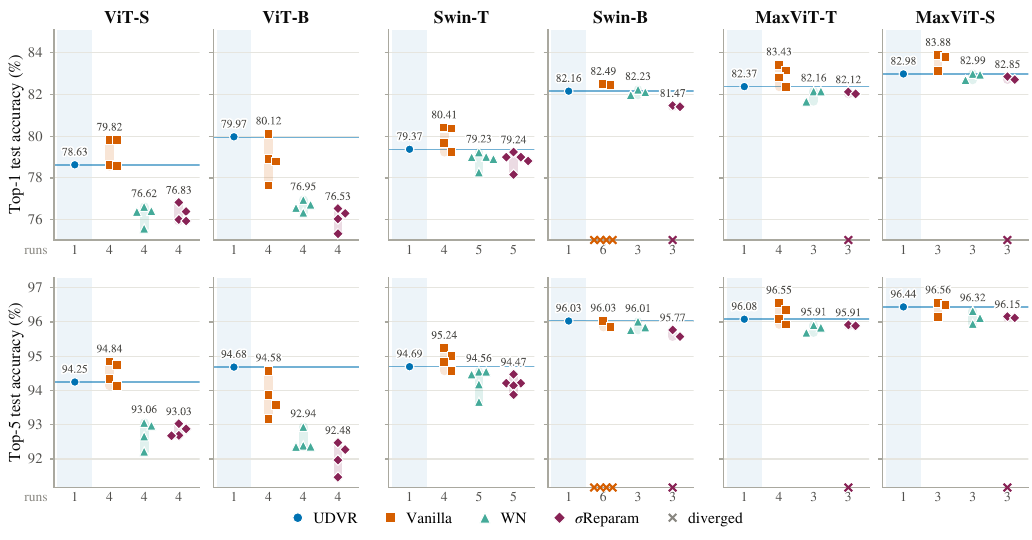}
  \caption{\textbf{Baseline sweeps on ImageNet.}
  Top-1 (top) and top-5 (bottom) accuracy of every run in the sweep of each baseline, compared with a single run of tuning-free UDVR.}
  \label{fig:baseline-imagenet}
\end{figure*}

\para{ImageNet results.}
\Cref{fig:baseline-imagenet} extends the comparison to ImageNet with two sizes each of ViT, Swin, and MaxViT.
UDVR uses a single run with the $C$ selected on CIFAR, whereas each baseline is swept over several runs on ImageNet.
Our method matches or exceeds WN and $\sigma$Reparam on every architecture in both top-1 and top-5 accuracy, with a clear margin on ViT.
The best swept Vanilla run leads UDVR by at most about one point in top-1 accuracy, and the gap narrows further in top-5, where UDVR ties Vanilla on Swin-B and surpasses it on ViT-B.
Moreover, several runs of Vanilla and $\sigma$Reparam diverge on Swin and MaxViT, whereas UDVR trains stably on all architectures.

\subsection{Language task}
\label{sec: llm}

\para{Setup.}
We pretrain GPT-2-style \citep{radford2019language, karpathy2022nanogpt}
models of the three sizes listed in \Cref{tab:models} on the same subset of SlimPajama \citep{cerebras2023slimpajama}.
The subset contains 1B tokens from each of 7 domains (arxiv, book, c4, commoncrawl, github, stackexchange, and wikipedia), giving 7B tokens in total, and is consumed in a single pass without repetition.
A held-out validation set of 15M tokens per domain (105M tokens in total) is used for evaluation.
We train with AdamW, a global batch of 128 sequences of length 2048, and a learning rate schedule with linear warmup over the first 1\% of steps followed by cosine decay to zero.
All experiments are repeated with 3 seeds, and we report validation perplexity averaged over the last 5 evaluation points and across seeds.
For the baselines, we sweep the learning rate at the small and medium sizes over a wide grid, together with the weight decay for Vanilla.
At the large size, the grid is narrowed around the optimum found at the smaller sizes.
In contrast, UDVR uses the same $C$ as in the vision experiments at all three model sizes, with no search or re-tuning.

\begin{figure*}[htb!]
  \centering
  \includegraphics[width=\textwidth]{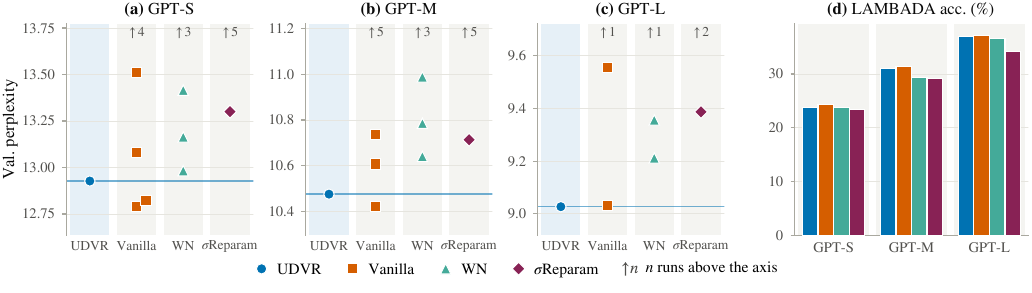}
  \caption{\textbf{LLM pretraining.}
  (a), (b), and (c) Validation perplexity of every sweep run (lower is better), with the horizontal line marking UDVR.
  (d) LAMBADA accuracy (higher is better), evaluated with the best sweep run of each baseline and with the single UDVR run.}
  \label{fig:llm-gpt2}
\end{figure*}

\para{Language modeling.}
\Cref{fig:llm-gpt2} reports the validation perplexity of GPT-2-style models at three sizes.
Without any search, UDVR outperforms the best WN and $\sigma$Reparam runs at every size and closely matches the best Vanilla run found by the sweep.
Its gap to Vanilla narrows as the model grows and disappears at GPT-L.
Most of the other baseline runs are considerably worse, indicating that the baselines are sensitive to the choice of hyperparameters.
On LAMBADA, all methods are close.
Vanilla is slightly ahead at all three sizes, and UDVR is ahead of both reparameterization baselines throughout.

\section{Discussion}
\label{sec: discussion}

\para{Robust is not always optimal.}
The deployment rule fixes one constant $C$ for every model and dataset, and it is chosen for robustness rather than for the optimum of any single model.
Its value lies in removing the sweep on a new model, together with the risk that such a sweep settles on a poor configuration.
This choice has a measurable cost.
On ImageNet, UDVR trails the grid-tuned Vanilla baseline by a small margin across ViT, Swin, and MaxViT, while it matches or exceeds both reparameterization baselines on every model.
On CIFAR, UDVR becomes more sensitive to large values of $C$ at $d=256$, where the largest $C$ in \Cref{fig:cifar_hyper_robust} causes a sharp accuracy drop.
To test whether this reflects a limitation of UDVR, we extend the sweep of $C$ to wider ViTs in \Cref{app:cifar_widedeep}..
The optimal learning rate of the tuned Vanilla baseline also shrinks with width, and its large learning rates collapse on wide models in the same way.
This indicates that the narrowing range of stable step sizes is a property of wide ViTs in this CIFAR setting rather than of our method.

\para{Scope of the reparameterization.}
UDVR acts only on the $(U,V)$ pairs in the attention and feed-forward blocks, as the diagonal $D$ requires two consecutive linear maps.
The fraction of parameters it covers therefore depends on the architecture.
In ViT and Swin, the large majority of parameters lie in such pairs.
In MaxViT, the convolutions of the MBConv blocks account for a substantial share of the parameters and remain outside UDVR (\Cref{tab:app-mainwd}).
On ImageNet, the gap of UDVR to the tuned Vanilla baseline shrinks from the smaller to the larger model in ViT and Swin, whereas it stays nearly constant in MaxViT.
The share of parameters outside UDVR may partly explain this difference.
Extending UDVR to a single weight matrix by factorizing it as $W = UDV$ remains an open question.
Such a factorization adds linear layers to the network and alters its optimization dynamics, although the factors can be merged into one matrix after training.

\para{Ablation study.}
\Cref{app:addresult} reports additional ablations and further discussion.
These cover wider and deeper models (\Cref{app:cifar_widedeep}), UDVR with weight decay on the non-UDV parameters (\Cref{app:udvr_wd}), the SwiGLU activation (\Cref{app:swiglu}), a comparison with $\mu$P on hyperparameter transfer across models and datasets (\Cref{app:mup}), and the transfer of the predefined $C$ to other LLaMA-style and MoE language models (\Cref{app:transferC_morellm}).

\section{Conclusion}
\label{sec: conclusion}

We presented UDV-Rebalance (UDVR), a reparameterization that inserts a diagonal scale buffer between the natural $(U, V)$ pairs of Transformer models.
UDVR rebalances each column of $U$ and each row of $V$ to a fixed norm.
It comes with a deployment rule that fixes the training hyperparameters in advance from the layer shape and a single scale constant $C$.

Constraining $U$ and $V$ to a fixed, data-independent norm at every step removes weight scale as a variable that training can adjust.
What remains is a geometric relationship between a parameter-space update and its effect on the composed map.
This relationship is determined by the layer shape rather than by the trajectory of training.
We view this as the underlying reason why a shape-only rule for the learning rate is possible in the first place.
Unpaired dense layers and normalization schemes that keep scale as a trainable parameter have no equivalent fixed reference.
Their learning rate therefore has to be found empirically for each model.

The UDVR deployment rule fit on vision transfers directly to language pretraining.
Every model size in that domain then deploys at a single predicted point with no further sweep.
The resulting zero-sweep configuration matches or exceeds both reparameterization baselines on both vision and language tasks, although it does not fully close the gap to the swept Vanilla baseline.
This is evidence that the shape-only criterion transfers across the architectures and scales we study and from vision to language, rather than being an artifact of the vision setting it was first fit to.

More broadly, a hyperparameter rule derived from layer shape turns a per-model search into a one-time design decision.
It removes a cost that would otherwise recur with each new architecture or scale a practitioner wants to try.
We expect the same principle, fixing a norm and pushing scale into a decoupled state variable updated by a rebalance rule, to extend beyond the learning rate to other training hyperparameters and to other reparameterizations that share this structure.

\section*{Acknowledgments}
This work is supported by the Wallenberg AI, Autonomous Systems, and Software Program (WASP), funded by the Knut and Alice Wallenberg Foundation.
Computational resources were provided on the Berzelius system funded by the Knut and Alice Wallenberg Foundation and operated by NAISS.

\bibliography{references}
\bibliographystyle{plainnat}

\clearpage
\newpage
\appendix

\section{Rebalance and Convergence}
\label{app:proofs}

This section records basic convergence behavior for an idealized GD/SGD version of UDV-Rebalance. 
The main point is that the rebalance itself does not obstruct the standard descent argument when used for networks with positively homogeneous activations, meaning activations satisfying 
$$\phi(cz)=c\phi(z)\quad\text{for all inputs } z \text{ and all } c>0.$$
Common activations like ReLU and leaky ReLU satisfy this condition.

\para{Rebalance map.}
Given $\widetilde{U}\in\mathbb{R}^{d\times r}$, $D\in\mathrm{diag}(\mathbb{R}^r)$, and $\widetilde{V}\in\mathbb{R}^{r\times d}$, assume that all columns of $\widetilde{U}$ and all rows of $\widetilde{V}$ are nonzero.
If this condition fails in an implementation, one can add a small constant $\epsilon$ to the denominators to avoid $0/0$. 
Define $S=\mathrm{Diag}(s)$ and $R=\mathrm{Diag}(\rho)$, such that
$$
s_k = \frac{\|\widetilde{U}_{:,k}\|_2}{{\alpha\sqrt{d}}}, \quad\text{and}\quad
\rho_k = \frac{\|\widetilde{V}_{k,:}\|_2}{{\alpha\sqrt{d}}},
$$
and set
$$
U^+=\widetilde{U} S^{-1}, \quad
V^+=R^{-1}\widetilde{V}, \quad\text{and}\quad
D^+=S D R.
$$
Then
$
\|U^+_{:,k}\|_2={\alpha\sqrt{d}}$ and $ 
\|V^+_{k,:}\|_2={\alpha\sqrt{d}}$ for all $k$, hence
$$ 
U^+D^+V^+ = \widetilde{U} S^{-1}(S D R)R^{-1}\widetilde{V} =\widetilde{U} D\widetilde{V}.
$$
Thus the rebalance pins the factor norms while leaving the composed $UDV$ product unchanged. 

The same rescaling also preserves an FFN block of the form $H(X;U,D,V)=\phi(XUD)V$, provided that $\phi$ is positively homogeneous. 
Since the rebalance scalings $S$ and $R$ have positive diagonal entries,
$$
\begin{aligned}
H(X;U^+,D^+,V^+)
&=\phi(X\widetilde{U} S^{-1}S D R)R^{-1}\widetilde{V}\\
&=\phi(X\widetilde{U} D R)R^{-1}\widetilde{V}\\
&=\phi(X\widetilde{U} D)RR^{-1}\widetilde{V}\\
&=\phi(X\widetilde{U} D)\widetilde{V}.
\end{aligned}
$$

\subsection{Deterministic gradient descent.}
For simplicity, consider the idealized objective
$$
F(U,D,V)=f(UDV),
$$
where $f$ is lower bounded. 
At iteration $t$, keep $D_t$ fixed during the gradient step and define
$$
\Phi_t(U,V)=f(UD_tV).
$$
Assume that, for all iterates under consideration, $\Phi_t$ has an $L$-Lipschitz gradient in the pair $(U,V)$ on the line segment between the current point and the provisional gradient step. 
With a common stepsize
$\eta\leq 1/L$, set
$$
\begin{aligned}
\widetilde{U}_{t+1}
& = U_t-\eta\nabla_U\Phi_t(U_t,V_t),\\
\widetilde{V}_{t+1} & = V_t-\eta\nabla_V\Phi_t(U_t,V_t),
\end{aligned}
$$
and then rebalance
$$
(U_{t+1},D_{t+1},V_{t+1}) = \mathrm{Rebalance}(\widetilde{U}_{t+1},D_t,\widetilde{V}_{t+1}).
$$
Let $G_t^2 = \|\nabla_U\Phi_t(U_t,V_t)\|_F^2 + \|\nabla_V\Phi_t(U_t,V_t)\|_F^2$. 
By the descent lemma applied to $\Phi_t$,
$$
\begin{aligned}
\Phi_t(\widetilde{U}_{t+1},\widetilde{V}_{t+1})
& \leq  \Phi_t(U_t,V_t) - \eta G_t^2 + \frac{L\eta^2}{2} G_t^2\\
& \leq \Phi_t(U_t,V_t)-\frac{\eta}{2} G_t^2.
\end{aligned}
$$
The rebalance identity gives
$$
\begin{aligned}
F(U_{t+1},D_{t+1},V_{t+1})
& = \Phi_t(\widetilde{U}_{t+1},\widetilde{V}_{t+1}),\\
F(U_t,D_t,V_t) & =\Phi_t(U_t,V_t).
\end{aligned}
$$
Therefore, we get
$$
F(U_{t+1},D_{t+1},V_{t+1}) \leq F(U_t,D_t,V_t)-\frac{\eta}{2}G_t^2.
$$
By summing from $t=0$ to $T-1$, and writing $F_{\inf}$ for any lower bound on $F$, we obtain
$$
\sum_{t=0}^{T-1}G_t^2
\le
\frac{2(F(U_0,D_0,V_0)-F_{\inf})}{\eta},
$$
and hence
$$
\min_{0\leq t<T}G_t^2
\le
\frac{2(F(U_0,D_0,V_0)-F_{\inf})}{\eta T}.
$$
This is the standard nonconvex $O(1/T)$ stationarity guarantee for the variables updated by gradient descent, with the rebalance step inserted after each update.

\subsection{Stochastic gradient descent.}
The same argument gives the usual SGD bound.
\begin{proposition}[Convergence of rebalanced SGD]\label{prop:convergence_rate}
Suppose that $F(U,D,V)$ is bounded below by $F_{\inf}$ and that
the objective $\Phi_t(U,V)$, with $D_t$ held fixed, has
$L$-Lipschitz continuous gradients. Let
\[
G_t^2 =
\|\nabla_U\Phi_t(U_t,V_t)\|_F^2
+
\|\nabla_V\Phi_t(U_t,V_t)\|_F^2.
\]
Consider the stochastic gradient updates
\[
\begin{aligned}
\widetilde U_{t+1} &= U_t-\eta g_{U,t},\\
\widetilde V_{t+1} &= V_t-\eta g_{V,t},
\end{aligned}
\]
followed by the rebalancing step, which preserves the
objective value:
\[
F(U_{t+1},D_{t+1},V_{t+1})
=
F(\widetilde U_{t+1},D_t,\widetilde V_{t+1}).
\]
Suppose that $g_t=(g_{U,t},g_{V,t})$ is conditionally unbiased
and has bounded conditional variance:
\begin{equation}
\label{eqn:stochastic-oracle-assumption}
\begin{aligned}
\mathbb{E}[g_t\mid\mathcal F_t]=\nabla\Phi_t(U_t,V_t),
\qquad\text{and}\qquad
\mathbb{E}\left[
\|g_t-\nabla\Phi_t(U_t,V_t)\|^2 \mid \mathcal F_t \right]\leq \sigma^2,
\end{aligned}
\end{equation}
where $\mathcal{F}_t$ denote the training history before sampling the step-$t$ stochastic gradient, so that $(U_t,D_t,V_t)$ is $\mathcal F_t$-measurable. 
Then, for any constant stepsize $\eta\leq 1/L$,
\[
\frac{1}{T}\sum_{t=0}^{T-1}\mathbb{E}[G_t^2]
\leq
\frac{2(F(U_0,D_0,V_0)-F_{\inf})}{\eta T}
+L\eta\sigma^2.
\]
In particular, choosing $\eta=\Theta(T^{-1/2})$ gives
\[
\frac{1}{T}\sum_{t=0}^{T-1}\mathbb{E}[G_t^2]
=O(T^{-1/2}).
\]
\end{proposition}

\begin{proof}

Let $g_t=(g_{U,t},g_{V,t})$ be an unbiased stochastic gradient of $\Phi_t$ at $(U_t,V_t)$, conditional on the past, and assume \eqref{eqn:stochastic-oracle-assumption} holds.
The provisional step is
$$
\begin{aligned}
\widetilde{U}_{t+1} & = U_t-\eta g_{U,t}, \\
\widetilde{V}_{t+1} & = V_t-\eta g_{V,t},
\end{aligned}
$$
followed by the same rebalance step. 
For $\eta \leq 1/L$, conditioning on
$\mathcal{F}_t$ gives
$$
\mathbb{E}[F(U_{t+1},D_{t+1},V_{t+1})\mid\mathcal{F}_t] 
\leq F(U_t,D_t,V_t) - \frac{\eta}{2}G_t^2 + \frac{L\eta^2}{2}\sigma^2. 
$$
After summing and taking expectations,
$$
\frac{1}{T}\sum_{t=0}^{T-1}\mathbb{E}[G_t^2] 
\leq \frac{2(F(U_0,D_0,V_0)-F_{\inf})}{\eta T} + L\eta\sigma^2.
$$
Thus a constant stepsize converges to the standard stochastic neighborhood, while choosing $\eta=\Theta(T^{-1/2})$ gives the usual $O(T^{-1/2})$ averaged stationarity rate.
\end{proof}

\section{Factor Pairs in Transformer Blocks}
\label{app:udv_pairs}

Following \citet{elhage2021mathematical}, multi-head attention with $h$ heads can be written as
\begin{align*}
\mathrm{MultiHead}(X)
= \mathrm{Concat}(O_1, \dots, O_h)\, W^O 
&= \sum_{i=1}^h O_i W_o^i
= \sum_{i=1}^h
\mathrm{softmax}\!\left(
\frac{Q_i K_i^\top}{\sqrt{d_{\text{head}}}}
\right) V_i W_o^i&\\
&= \sum_{i=1}^h
\mathrm{softmax}\!\left(
\frac{
X \bigl(W_q^i {W_k^i}^\top\bigr) X^\top
}{\sqrt{d_{\text{head}}}}
\right)
X \bigl(W_v^i W_o^i\bigr),&
\end{align*}
where $Q_i = XW_q^i$, $K_i = XW_k^i$, and $V_i = XW_v^i$ are the per-head query, key, and value projections, and $W_o^i$ is the row-block of $W^O$ that corresponds to head $i$.
Biases are omitted, and $V_i$ denotes the value projection, not the factor $V$.
The weights of head $i$ therefore enter the output only through the products $W_q^i {W_k^i}^\top$ and $W_v^i W_o^i$.
This gives the factor pairs $(W_q^i,{W_k^i}^\top)$ and $(W_v^i,W_o^i)$, which play the roles of $(U,V)$ in our construction.
The FFN has the analogous form
\begin{equation*}
\mathrm{FFN}(X) = \phi\!\left(X W_{\text{up}}\right) W_{\text{down}},
\end{equation*}
with $W_{\text{up}}$ as $U$, $W_{\text{down}}$ as $V$, and $\phi$ an activation function.

\section{Pseudocode of UDVRebalance}
\label{app:pseudocode}

\begin{algorithm}[H]
\caption{\textsc{Rebalance}$(U,D,V)$}
\label{alg:rebalance}
\begin{algorithmic}[1]
\REQUIRE factors $U\in\mathbb{R}^{d\times r}$ and $V\in\mathbb{R}^{r\times d}$, diagonal buffer $D\in\mathbb{R}^{r\times r}$, norm scaler $\alpha$
\ENSURE every column of $U$ and every row of $V$ pinned to norm $\alpha\sqrt{d}$
\FOR{$k = 1$ \TO $r$}
    \STATE $s_k \leftarrow \lVert U_{:,k}\rVert_2 \,/\, (\alpha\sqrt{d})$ 
    \STATE $\rho_k \leftarrow \lVert V_{k,:}\rVert_2 \,/\, (\alpha\sqrt{d})$ 
\ENDFOR
\STATE $S \leftarrow \mathrm{Diag}(s),\qquad R \leftarrow \mathrm{Diag}(\rho)$
\STATE $U \leftarrow U S^{-1}$ 
\STATE $V \leftarrow R^{-1} V$ 
\STATE $D \leftarrow S\,D\,R$ 
\STATE \textbf{return} $(U,D,V)$
\end{algorithmic}
\end{algorithm}

\begin{algorithm}[H]
\caption{UDVRebalance training of one reparameterized pair}
\label{alg:udvrebalance}
\begin{algorithmic}[1]
\REQUIRE initial factors $U\in\mathbb{R}^{d\times r}$ and $V\in\mathbb{R}^{r\times d}$, norm scaler $\alpha$, learning rate $\eta$ from \Cref{eq:width-invariant}, regularizer scaler $\gamma$, task loss $f$
\ENSURE trained $(U,D,V)$ with every column of $U$ and every row of $V$ pinned to norm $\alpha\sqrt{d}$ after each step
\STATE $D \leftarrow I_r$ 
\STATE $(U,D,V) \leftarrow \textsc{Rebalance}(U,D,V)$ 
\COMMENT{\Cref{alg:rebalance}}
\FOR{each training step}
    \STATE $\mathcal{L} \leftarrow f + \gamma\,\bigl(\lVert UD\rVert_F^2 + \lVert DV\rVert_F^2\bigr)$ 
    \COMMENT{$\mathcal{R}_{\text{attn}}$ in \Cref{eq:regularizer}}
    \STATE $G_U \leftarrow \nabla_U \mathcal{L},\qquad G_V \leftarrow \nabla_V \mathcal{L}$ 
    \STATE $U \leftarrow \textsc{AdamStep}(U, G_U, \eta)$ 
    \STATE $V \leftarrow \textsc{AdamStep}(V, G_V, \eta)$
    \STATE $(U,D,V) \leftarrow \textsc{Rebalance}(U,D,V)$ 
\ENDFOR
\end{algorithmic}
\end{algorithm}

Each reparameterized pair consists of two existing linear maps $U\in\mathbb{R}^{d\times r}$ and $V\in\mathbb{R}^{r\times d}$ from the original model (the query and key projections or the value and output projections of an attention head, or the two linear layers of a feed-forward block), between which we insert a diagonal scale buffer $D\in\mathbb{R}^{r\times r}$ kept out of the optimizer.
Here $r$ is the inner dimension shared by the two factors, the native width of the operator.
A single coefficient $\alpha$ fixes the per-column / per-row target norm $\alpha\sqrt{d}$, so that after every rebalance $\lVert U_{:,k}\rVert_2=\lVert V_{k,:}\rVert_2=\alpha\sqrt{d}$ for all $k$.
Both factors of a pair train at the same rate $\eta$ from \Cref{eq:width-invariant}.
For a pair inside an attention head, $\eta$ is $\eta_{\text{attn}}$, computed with the factors of all heads stacked together.

\para{Layers with an activation.}
Since $D$ lies between two linear maps in attention, rebalancing preserves $UDV$.
In a feed-forward block, the activation $\phi$ separates the two factors, and the block maps a row vector $x$ to $\phi(xUD)\,V$, with $D$ placed between $U$ and $\phi$.
The scale stripped from $U$ still passes into $D$ without changing the output, whereas the scale stripped from $V$ would have to cross $\phi$, which is uncommon for activations beyond the positively homogeneous ones.
For a feed-forward pair, one line therefore changes in each algorithm.
In \Cref{alg:rebalance}, the update of $D$ becomes $D \leftarrow S\,D$ and the stripped scale of $V$ is discarded, while each row of $V$ is still pinned to $\alpha\sqrt{d}$.
Since the rebalance runs after every step, each row of $V$ stays on the sphere of radius $\alpha\sqrt{d}$, and the output scale of the block is carried by $D$ through the $U$ side alone.
In \Cref{alg:udvrebalance}, the loss uses $\mathcal{R}_{\text{ffn}}$ in place of $\mathcal{R}_{\text{attn}}$.
Under $\mathcal{R}_{\text{attn}}$, the shrinkage that the $DV$ term applies to $V$ would be discarded, and only the $\gamma\lVert UD\rVert_F^2$ part would act on $D$.
On the sphere this part equals $\gamma\alpha^{2}d\,\lVert D\rVert_F^2$, half of the $2\gamma\alpha^{2}d\,\lVert D\rVert_F^2$ that attention applies.
This is why $\mathcal{R}_{\text{ffn}}$ carries a factor of $2$, which doubles the $U$ part to $2\gamma\lVert UD\rVert_F^2$ and restores the full penalty on $D$.

\section{Additional Experimental Details}
\label{app:exp_details}

\subsection{Computational resource.}
All experiments are conducted on NVIDIA A100 (40GB) GPUs. 

\subsection{Vision pretraining setting.}
Our vision experiments are conducted on CIFAR-10/100, DomainNet and ImageNet, well-known benchmarks in the vision literature.
\Cref{app:tab model_configs} lists the ViT scales used in our computer vision experiments, whose architecture design follows the DeiT/ViT recipe.
\Cref{app:tab exp_setup} lists the training-related settings and hyperparameters shared across all four compared methods (Vanilla, WN, $\sigma$Reparam, UDVR) on vision tasks.
This protocol (architecture, data, epochs, augmentation, and seeds) is held identical across methods, so that any accuracy difference is attributable to the reparameterization method itself.
To obtain robust results, all experiments are conducted with three random seeds and reported metrics are averaged across the three seeds, except the experiments on ImageNet.

\begin{table}[htb!]
\centering
\caption{Model configurations across datasets.}
\resizebox{\columnwidth}{!}{%
\begin{tabular}{lccccccrrr}
\toprule
Model & Depth & Heads & $d$ & $d_{\text{head}}$ & $d_{\text{mlp}}$ & Drop path
      & \multicolumn{3}{c}{Trainable params (M)} \\
\cmidrule(lr){8-10}
\multicolumn{7}{l}{\textit{CIFAR ($32^2$, patch 4) / DomainNet ($64^2$, patch 8)}}
      & CIFAR-10 & CIFAR-100 & DomainNet \\
\midrule
ViT & 6  & 2  & 128  & 64 & 512  & 0.0 & 1.20   & 1.21   & 1.26  \\
ViT & 8  & 2  & 128  & 64 & 512  & 0.0 & 1.59   & 1.60   & 1.65  \\
ViT & 12 & 2  & 128  & 64 & 512  & 0.1 & 2.38   & 2.39   & 2.44  \\
\cmidrule(lr){1-10}
ViT & 6  & 3  & 192  & 64 & 768  & 0.0 & 2.68   & 2.70   & 2.77  \\
ViT & 8  & 3  & 192  & 64 & 768  & 0.0 & 3.57   & 3.59   & 3.66  \\
ViT & 12 & 3  & 192  & 64 & 768  & 0.1 & 5.34   & 5.36   & 5.43  \\
\cmidrule(lr){1-10}
ViT & 6  & 4  & 256  & 64 & 1024 & 0.0 & 4.76   & 4.78   & 4.88  \\
ViT & 8  & 4  & 256  & 64 & 1024 & 0.0 & 6.33   & 6.35   & 6.45  \\
ViT & 12 & 4  & 256  & 64 & 1024 & 0.1 & 9.48   & 9.50   & 9.60  \\
\midrule
\multicolumn{7}{l}{\textit{ImageNet-1k ($224^2$)}} & \multicolumn{3}{c}{ImageNet} \\
\midrule
ViT-S    & 12       & 6         & 384       & 64 & 1536 & 0.1 & \multicolumn{3}{c}{22.01} \\
ViT-B    & 12       & 12        & 768       & 64 & 3072 & 0.2 & \multicolumn{3}{c}{86.48} \\
Swin-T   & 2-2-6-2  & 3-6-12-24 & 96--768   & 32 & $4d$ & 0.2 & \multicolumn{3}{c}{28.25} \\
Swin-B   & 2-2-18-2 & 4-8-16-32 & 128--1024 & 32 & $4d$ & 0.5 & \multicolumn{3}{c}{87.66} \\
MaxViT-T & 2-2-5-2  & 2-4-8-16  & 64--512   & 32 & $4d$ & 0.2 & \multicolumn{3}{c}{30.60} \\
MaxViT-S & 2-2-5-2  & 3-6-12-24 & 96--768   & 32 & $4d$ & 0.3 & \multicolumn{3}{c}{68.26} \\
\bottomrule
\end{tabular}%
}
\label{app:tab model_configs}
\end{table}

\begin{table}[htbp!]
\centering
\caption{Experimental setup and common hyperparameters shared across all compared methods on CIFAR-10/CIFAR-100, DomainNet and ImageNet.}
\renewcommand{\arraystretch}{1.15}
\resizebox{\textwidth}{!}{%
\begin{tabular}{@{}ll p{4.6cm} p{4.6cm} p{4.6cm}@{}}
\toprule
\textbf{Category} & \textbf{Item} & \textbf{CIFAR-10 / CIFAR-100} & \textbf{DomainNet (real / sketch)} & \textbf{ImageNet} \\
\midrule
\multirow{4}{*}{Optimizer}
 & Optimizer & \multicolumn{3}{c}{AdamW} \\
 & Betas & \multicolumn{3}{c}{(0.9, 0.999)} \\
 & Eps & \multicolumn{3}{c}{1e-8} \\
 & Gradient clip & \multicolumn{3}{c}{None} \\
\midrule
\multirow{2}{*}{LR schedule}
 & Scheduler & \multicolumn{3}{p{14.4cm}}{\centering linear warmup then cosine decay to 0, stepped per optimizer step} \\
 & Warmup ratio & \multicolumn{3}{p{14.4cm}}{\centering 0.05} \\
\midrule
\multirow{4}{*}{Image}
 & Classes & 10 / 100 & 345 & 1000 \\
 & Preprocessing & -- & official split, whole image resized offline to $64 \times 64$ (box filter, no crop) & -- \\
 & Image size & 32 & 64 & 224 \\
 & Patch size & 4 & 8 & 16 \\
\midrule
\multirow{2}{*}{Normalization}
 & Mean & CIFAR-10 (0.4914, 0.4822, 0.4465)\newline CIFAR-100 (0.5071, 0.4867, 0.4408)
        & real (0.6066, 0.5898, 0.5564)\newline sketch (0.8326, 0.8269, 0.8180)
        & (0.485, 0.456, 0.406) \\
 & Std  & CIFAR-10 (0.2023, 0.1994, 0.2010)\newline CIFAR-100 (0.2675, 0.2565, 0.2761)
        & real (0.3220, 0.3153, 0.3377)\newline sketch (0.2419, 0.2444, 0.2509)
        & (0.229, 0.224, 0.225) \\
\midrule
Augmentation & Train & AutoAugment \citep{cubuk2018autoaugment}
                     & AutoAugment \citep{cubuk2018autoaugment}
                     & RandomResizedCrop(224)\newline RandomHorizontalFlip\newline AutoAugment \citep{cubuk2018autoaugment} \\
\midrule
\multirow{4}{*}{Mixup / Loss}
 & mixup\_alpha & \multicolumn{3}{c}{0.8} \\
 & cutmix\_alpha & \multicolumn{3}{c}{1.0} \\
 & mix\_prob / switch\_prob & \multicolumn{3}{c}{0.8 / 0.5} \\
 & Label smoothing & \multicolumn{3}{c}{0.1} \\
\midrule
\multirow{5}{*}{Training}
 & Precision & \multicolumn{3}{c}{Autocast bfloat16} \\
 & Batch size & 512, single GPU & 512, single GPU & global 2048 = 256 per-rank $\times$ 8 GPUs \\
 & Num epochs & CIFAR-10: 200\newline CIFAR-100: 300
              & real: 123\newline sketch: 310\newline (matches the CIFAR-100 step count, $\approx$28.5k steps)
              & 300 \\
 & Seed & \multicolumn{3}{c}{42 (the other two random seeds are derived from this seed; ImagetNet with 42 only)} \\
 & Data split & \multicolumn{3}{c}{2\% of official train split held out as val, test uses official test split} \\
\bottomrule
\end{tabular}%
}
\label{app:tab exp_setup}
\end{table}

\Cref{app:tab hp_sweep_config,app:tab imagenet_baseline_grid} lists the sweep grid for each method.
Method-specific hyperparameters introduced by each baseline follow the configuration in its original paper, while common hyperparameters are swept.
Vanilla receives a full learning-rate $\times$ weight-decay grid.
Weight Normalization (WN) and $\sigma$Reparam ($\sigma$R) disable weight decay following their reference implementations, so only the learning rate is swept.
The sweep on UDVR with CIFAR-10 and CIFAR-100 yields the tuning-free deployment rule used in the main comparison and is not part of UDVR's deployment cost.

\begin{table}[htb!]
\centering
\caption{Hyperparameter sweep configurations and method-specific implementation details for the vision experiments.}
\small
\setlength{\tabcolsep}{4pt}
\renewcommand{\arraystretch}{1.15}
\newcolumntype{L}{>{\raggedright\arraybackslash}X}
\begin{tabularx}{\textwidth}{@{}>{\raggedright\arraybackslash}p{3.1cm}LLLL@{}}
\toprule
Hyperparameter & Vanilla & WN & $\sigma$Reparam & UDVR (predefined) \\
\midrule
$\eta_{\text{main}}$ &
$10^{-4}$, $3{\times}10^{-4}$, $10^{-3}$, $3{\times}10^{-3}$, $10^{-2}$ &
$10^{-5}$, $3{\times}10^{-5}$, $10^{-4}$, $3{\times}10^{-4}$, $10^{-3}$, $3{\times}10^{-3}$, $10^{-2}$, $3{\times}10^{-2}$, $10^{-1}$ &
$10^{-5}$, $3{\times}10^{-5}$, $10^{-4}$, $3{\times}10^{-4}$, $10^{-3}$, $3{\times}10^{-3}$, $10^{-2}$, $3{\times}10^{-2}$, $10^{-1}$ &
$3{\times}10^{-4}$ (CIFAR, DomainNet; batch 512) \newline
$1.2{\times}10^{-3}$ (ImageNet; batch 2048) \\
\addlinespace
Weight decay &
$0$, $10^{-4}$, $10^{-3}$, $10^{-2}$, $0.05$, $0.1$ &
$0$ & $0$ & $0$ \\
\addlinespace
Runs per cell (CIFAR) & 30 & 9 & 9 & 1 \\

\midrule
\multicolumn{5}{@{}l}{\textit{WN}} \\
Weight form & -- & $w = (g/\lVert\hat{w}\rVert)\,\hat{w}$ & -- & -- \\
\midrule
\multicolumn{5}{@{}l}{\textit{$\sigma$Reparam}} \\
Spectral scale $\zeta$ & -- & -- &
SVD-initialized to $s[0]$, the largest singular value of $W$ & -- \\
Power iteration & -- & -- & 1 step per training forward & -- \\
$u$ buffers & -- & -- & Single buffer & -- \\
Spectral precision & -- & -- &
fp32; main matmuls in shared bf16 autocast & -- \\
\midrule
\multicolumn{5}{@{}l}{\textit{UDVR}} \\
$\alpha$ (pinning scale) & -- & -- & -- & $10^{-2}$ \\
$C$ (LR constant) & -- & -- & -- &
$0.0217$ \\
$\eta_{\text{attn}}$ (derived) & -- & -- & -- & $C\alpha = 2.17{\times}10^{-4}$ \\
$\eta_{\text{ffn}}$ (derived) & -- & -- & -- & $2C\alpha = 4.34{\times}10^{-4}$ \\
$\gamma$ (regularizer) & -- & -- & -- & $10^{-7}$ \\
\bottomrule
\end{tabularx}
\label{app:tab hp_sweep_config}
\end{table}

\begin{table}[htb!]
\centering
\caption{Per-architecture baseline sweep grids on ImageNet (seed 42). The Vanilla grid is the full product of the listed $\eta_{\text{main}}$ and weight-decay values. WN and $\sigma$Reparam use weight decay $0$ throughout.}
\small
\setlength{\tabcolsep}{5pt}
\renewcommand{\arraystretch}{1.15}
\begin{tabularx}{\textwidth}{@{}l>{\raggedright\arraybackslash}X>{\raggedright\arraybackslash}p{1.9cm}>{\raggedright\arraybackslash}X>{\raggedright\arraybackslash}X@{}}
\toprule
& \multicolumn{2}{c}{Vanilla} & WN & $\sigma$Reparam \\
\cmidrule(lr){2-3}\cmidrule(lr){4-4}\cmidrule(l){5-5}
Model & $\eta_{\text{main}}$ & Weight decay & $\eta_{\text{main}}$ & $\eta_{\text{main}}$ \\
\midrule
ViT-S &
$10^{-3}$, $3{\times}10^{-3}$ & $0.05$, $0.1$ &
$3{\times}10^{-4}$, $10^{-3}$, $3{\times}10^{-3}$, $10^{-2}$ &
$3{\times}10^{-4}$, $10^{-3}$, $3{\times}10^{-3}$, $10^{-2}$ \\
ViT-B &
$10^{-3}$, $2{\times}10^{-3}$ & $0.05$, $0.1$ &
$3{\times}10^{-4}$, $10^{-3}$, $3{\times}10^{-3}$, $10^{-2}$ &
$3{\times}10^{-4}$, $10^{-3}$, $3{\times}10^{-3}$, $10^{-2}$ \\
\addlinespace
Swin-T &
$2{\times}10^{-3}$, $4{\times}10^{-3}$ & $0.05$, $0.1$ &
$3{\times}10^{-4}$, $10^{-3}$, $3{\times}10^{-3}$, $10^{-2}$, $3{\times}10^{-2}$ &
$3{\times}10^{-4}$, $10^{-3}$, $3{\times}10^{-3}$, $10^{-2}$, $3{\times}10^{-2}$ \\
Swin-B &
$10^{-3}$, $2{\times}10^{-3}$, $4{\times}10^{-3}$ & $0.05$, $0.1$ &
$10^{-3}$, $3{\times}10^{-3}$, $10^{-2}$ &
$10^{-2}$, $3{\times}10^{-2}$, $10^{-1}$ \\
\addlinespace
MaxViT-T &
$7.5{\times}10^{-4}$, $1.5{\times}10^{-3}$, $3{\times}10^{-3}$, $6{\times}10^{-3}$ & $0.05$ &
$10^{-3}$, $3{\times}10^{-3}$, $10^{-2}$ &
$10^{-2}$, $3{\times}10^{-2}$, $10^{-1}$ \\
MaxViT-S &
$1.5{\times}10^{-3}$, $3{\times}10^{-3}$, $6{\times}10^{-3}$ & $0.05$ &
$10^{-3}$, $3{\times}10^{-3}$, $10^{-2}$ &
$3{\times}10^{-3}$, $10^{-2}$, $3{\times}10^{-2}$ \\
\bottomrule
\end{tabularx}
\label{app:tab imagenet_baseline_grid}
\end{table}

\subsection{Language pretraining setting.}
Our language pretraining experiments are conducted on a subset of SlimPajama, a widely used large-scale corpus for LLM pretraining.
\Cref{app:tab llm_model_scales} lists the three scales of the GPT-2-style decoder-only Transformer used in our LLM pretraining experiments.
\Cref{app:tab llm_hyperparams} lists the training-related settings and hyperparameters shared across all four compared methods (Vanilla, WN, $\sigma$Reparam, UDVR) on language modeling tasks.
This protocol (architecture, tokenizer, data mixture, sequence length, batch size, schedule, and seeds) is held identical across methods, so that any perplexity difference is attributable to the reparameterization method itself.
Training follows a single pass over the token budget of 7 billion tokens, uniformly 1 billion tokens across each of the seven domains.
To obtain robust results, all experiments are conducted with three random seeds and reported metrics are averaged across the three seeds.

\begin{table}[h]
\centering
\caption{GPT model configurations across scales.}
\begin{tabular}{lcccccr}
\toprule
Model & Depth & Heads & $d$ & $d_{\text{head}}$ & $d_{\text{mlp}}$ & Trainable \\
\midrule
GPT-S & 22 & 8  & 512  & 64 & 2048 & 120.76M  \\
GPT-M & 22 & 16 & 1024 & 64 & 4096 & 379.94M  \\
GPT-L & 22 & 32 & 2048 & 64 & 8192 & 1313.53M \\
\bottomrule
\end{tabular}%
\label{app:tab llm_model_scales}
\end{table}

\Cref{app:tab hp_sweep_config_llm} lists the sweep grid for each method across the three scales.
As in the vision setting, method-specific hyperparameters introduced by each baseline follow the configuration in its original paper, while common hyperparameters are swept.
Vanilla receives a full learning-rate $\times$ weight-decay grid at the GPT-S (small) and GPT-M (mid) scales, narrowing to a learning-rate-only sweep at the GPT-L (large) scale.
WN and $\sigma$Reparam disable weight decay following their reference implementations, so only the learning rate is swept on a shared six-point grid at the small and mid scales, narrowing to three points at the large scale.
For UDVR, all hyperparameters such as $\alpha$, $C$, $\eta_{\text{main}}$, and $\gamma$ are fixed at the values transferred directly from the vision setting, requiring no additional per-scale tuning.

\begin{table}[htb]
\centering
\caption{LLM pre-training hyperparameters and configuration.}
\renewcommand{\arraystretch}{1.15}
\resizebox{\textwidth}{!}{%
\begin{tabular}{@{}lll@{}}
\toprule
\textbf{Category} & \textbf{Item} & \textbf{Value} \\
\midrule
\multirow{4}{*}{Optimizer}
 & Optimizer & AdamW \\
 & Betas & (0.9, 0.95) \\
 & Eps & 1e-8 \\
 & Gradient clip & None \\
\midrule
\multirow{2}{*}{LR schedule}
 & Scheduler & linear warmup then cosine decay to 0, stepped per optimizer step \\
 & Warmup ratio & 0.01 \\
\midrule
\multirow{8}{*}{Data}
 & Corpus & SlimPajama-627B \\
 & Tokenizer & GPT-NeoX BPE \\
 & Domains & arxiv, book, c4, commoncrawl, github, stackexchange, wikipedia \\
 & Sequence length & 2048 \\
 & Vocabulary size & 50,280 (padded to 50,304) \\
 & Global batch size & 128 sequences, constant across tiers \\
 & Per-tier (GPU, micro-batch, grad-accum) & Small (1, 16, 8), Mid (4, 8, 4), Large (8, 1, 16) \\
 & Token budget & 7e9 (uniform 1e9 $\times$ 7 domains, single pass) \\
 \midrule
 Seed & Random seed & 42 (the other two random seeds are derived from this seed) \\
 \midrule
 Precision & Numerical precision & Autocast bfloat16 \\
\midrule
\multirow{4}{*}{Architecture}
 & Positional encoding & RoPE \\
 & Normalization & LayerNorm (bias=True), pre-norm \\
 & FFN & Non-gated MLP, GELU activation \\
 & Embedding & Untied (Pythia-style) \\
\midrule
\multirow{2}{*}{Objective}
 & Train criterion & Next-token cross-entropy \\
 & Eval criterion & Same as train, aggregated per domain \\
\midrule
Evaluation & Eval subset size & 1.5e7 tokens per domain, unused in training \\
\bottomrule
\end{tabular}%
}
\label{app:tab llm_hyperparams}
\end{table}

\begin{table}[htb!]
\centering
\caption{Hyperparameter sweep configurations and method-specific implementation details for the LLM experiments, across the S/M/L scales.}
\small
\setlength{\tabcolsep}{4pt}
\renewcommand{\arraystretch}{1.15}
\newcolumntype{L}{>{\raggedright\arraybackslash}X}
\begin{tabularx}{\textwidth}{@{}>{\raggedright\arraybackslash}p{3.1cm}LLLL@{}}
\toprule
Hyperparameter & Vanilla & WN & $\sigma$Reparam & UDVR \\
\midrule
$\eta_{\text{main}}$, GPT-S &
$10^{-4}$, $3{\times}10^{-4}$, $10^{-3}$, $3{\times}10^{-3}$ &
$3{\times}10^{-5}$, $10^{-4}$, $3{\times}10^{-4}$, $10^{-3}$, $3{\times}10^{-3}$, $10^{-2}$ &
$3{\times}10^{-5}$, $10^{-4}$, $3{\times}10^{-4}$, $10^{-3}$, $3{\times}10^{-3}$, $10^{-2}$ &
$3{\times}10^{-4}$ \\
$\eta_{\text{main}}$, GPT-M &
$10^{-4}$, $3{\times}10^{-4}$, $10^{-3}$, $3{\times}10^{-3}$ &
$3{\times}10^{-5}$, $10^{-4}$, $3{\times}10^{-4}$, $10^{-3}$, $3{\times}10^{-3}$, $10^{-2}$ &
$3{\times}10^{-5}$, $10^{-4}$, $3{\times}10^{-4}$, $10^{-3}$, $3{\times}10^{-3}$, $10^{-2}$ &
$3{\times}10^{-4}$ \\
$\eta_{\text{main}}$, GPT-L &
$3{\times}10^{-4}$, $10^{-3}$, $3{\times}10^{-3}$ &
$3{\times}10^{-4}$, $10^{-3}$, $3{\times}10^{-3}$ &
$3{\times}10^{-4}$, $10^{-3}$, $3{\times}10^{-3}$ &
$3{\times}10^{-4}$ \\
\addlinespace
Weight decay &
$0$, $0.1$ (S, M) \newline $0.1$ (L) &
$0$ & $0$ & $0$ \\
\addlinespace
Runs (S / M / L) & 8 / 8 / 3 & 6 / 6 / 3 & 6 / 6 / 3 & 1 / 1 / 1 \\
\midrule
\multicolumn{5}{@{}l}{\textit{WN}} \\
Weight form & -- & $w = (g/\lVert\hat{w}\rVert)\,\hat{w}$ & -- & -- \\
\midrule
\multicolumn{5}{@{}l}{\textit{$\sigma$Reparam}} \\
Spectral scale $\zeta$ & -- & -- &
SVD-initialized to $s[0]$, the largest singular value of $W$ & -- \\
Power iteration & -- & -- & 1 step per optimizer step & -- \\
$u$ buffers & -- & -- & Single buffer & -- \\
Spectral precision & -- & -- &
fp32; main matmuls in shared bf16 autocast & -- \\
\midrule
\multicolumn{5}{@{}l}{\textit{UDVR}} \\
$\alpha$ (pinning scale) & -- & -- & -- & $10^{-2}$ \\
$C$ (LR constant) & -- & -- & -- & $0.0217$ (all scales) \\
$\eta_{\text{attn}}$ (derived) & -- & -- & -- & $C\alpha = 2.17{\times}10^{-4}$ \\
$\eta_{\text{ffn}}$ (derived) & -- & -- & -- & $2C\alpha = 4.34{\times}10^{-4}$ \\
$\gamma$ (regularizer) & -- & -- & -- & $10^{-7}$ \\
\bottomrule
\end{tabularx}
\label{app:tab hp_sweep_config_llm}
\end{table}

\section{UDVR Hyperparameter Selection}
\label{app:hparams}

The deployment rule in \Cref{sec: experiments} fixes three constants at $\alpha = 10^{-2}$, $\gamma = 10^{-7}$, and $C = 0.0217$.
This section explains how these values were obtained.
We first characterized an earlier variant whose step size was set by a scaler $\beta$ tuned for every model.
On that variant we fixed $\alpha$ and $\gamma$ and found that the best $\beta$ scales as $C/d$, which turned the per-model scaler into a single constant (\Cref{app:old-method}).
We then compared several forms of the FFN pair and kept the V projection (\Cref{app:ffn-variants}).
Finally, we selected $C$ for this form over a grid of widths and depths, with $\alpha$ and $\gamma$ carried over unchanged (\Cref{app:c-star}).

\subsection{Variants}
\label{app:variants}

All variants treat attention identically, with one diagonal per head and the rebalance of \eqref{eq:rebalance}.
They differ in how the FFN pair is rebalanced and in how the step size of $U$ and $V$ is set.

\paragraph{FFN forms.}
\Cref{tab:app-ffn-forms} lists the FFN forms we compare, which differ only in where the rebalance moves the stripped scales, and V-proj is the form used by UDVR.

\begin{table}[htb!]
\centering
\small
\caption{FFN forms compared in this ablation. The columns ``Scale of $U$'' and ``Scale of $V$'' state where the rebalance moves the scale stripped from each factor. $^\dagger$ indicates no norm constraint.}
\begin{tabular}{lccc}
\toprule
Form & FFN forward & Scale of $U$ & Scale of $V$ \\
\midrule
Single-D & $\phi(XUD)V$    & to $D$    & to $D$     \\
V-proj   & $\phi(XUD)V$    & to $D$    & discarded  \\
U-proj   & $\phi(XU)DV$    & discarded & to $D$     \\
U-only   & $\phi(XUD)V$    & to $D$    & discarded$^\dagger$ \\
Two-D    & $\phi(XUD)D_2V$ & to $D$    & to $D_2$   \\
\bottomrule
\end{tabular}
\label{tab:app-ffn-forms}
\end{table}

\paragraph{Step-size rules.}
UDVR sets the learning rate of each factor by \eqref{eq:lr-rule}, which we refer to as the split rule.
The earlier variant instead set the rate of each factor to a scaler $\beta$ times its pinned Frobenius norm, computed per head for attention factors,
\begin{equation}
\eta_{\text{attn}} = \beta\alpha\sqrt{d\,d_{\text{head}}},
\qquad
\eta_{\text{ffn}} = \beta\alpha\sqrt{d\,d_{\text{ffn}}}.
\label{eq:app-beta}
\end{equation}
We compare tuning $\beta$ for every model and dataset (per-model $\beta$) with the shared rule $\beta = C/d$, where a single constant $C$ serves all models.
With $d_{\text{ffn}} = 4d$ and $h = d/d_{\text{head}}$ heads, the shared rule gives $\eta_{\text{attn}} = C\alpha/\sqrt{h}$ and $\eta_{\text{ffn}} = 2C\alpha$, whereas the split rule gives $\eta_{\text{attn}} = C\alpha$ and the same $\eta_{\text{ffn}}$.
The two rules thus agree on the FFN and differ on attention by a factor of $\sqrt{h}$.

\subsection{From a per-model scaler to a single constant}
\label{app:old-method}

We swept Single-D with per-model $\beta$ over the grid in \Cref{tab:app-grid} on CIFAR-10 and CIFAR-100 with a single seed.
The sweep uses two ViTs without drop path, Nano with width $d = 128$ and depth 6, and Mini with $d = 192$ and depth 8, both with $d_{\text{head}} = 64$ and $d_{\text{ffn}} = 4d$.
Within a run, all UDV layers share the same $(\alpha, \beta, \gamma)$.

\begin{table}[htb!]
\centering
\small
\caption{Grid of the characterization sweep for Single-D with per-model $\beta$. The full grid is run for every model and dataset.}
\begin{tabular}{lcc}
\toprule
Hyperparameter & Values & Count \\
\midrule
$\alpha$             & $\{1, 3, 6\}{\times}10^{k}$ for $k \in \{-4, -3, -2\}$, $10^{-1}$, $1$ & 11 \\
$\beta$              & $\{1, 3, 6\}{\times}10^{k}$ for $k \in \{-5, -4\}$, $10^{-3}$         & 7  \\
$\gamma$             & $10^{-7}$, $10^{-5}$, $10^{-3}$                                        & 3  \\
$\eta_{\text{main}}$ & $3{\times}10^{-4}$, $10^{-3}$, $3{\times}10^{-3}$                      & 3  \\
\bottomrule
\end{tabular}%
\label{tab:app-grid}
\end{table}

\begin{figure}[htb!]
  \centering
  \includegraphics[width=\textwidth]{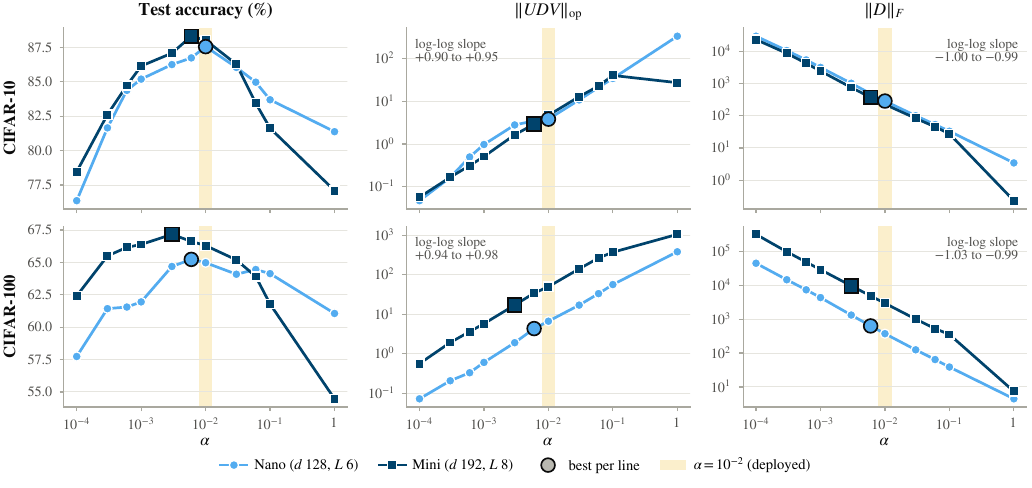}
  \caption{\textbf{Effect of the pinning scale.}
  Each line varies $\alpha$ in Single-D with $\beta$, $\gamma$, and $\eta_{\text{main}}$ fixed at the best configuration of each model and dataset.
  Left. Test accuracy (higher is better).
  Middle and right. Largest final $\lVert UDV\rVert_{\text{op}}$ and $\lVert D\rVert_F$ of the FFN pairs over all blocks.}
  \label{fig:app-alpha}
\end{figure}

\paragraph{Pinning scale.}
\label{app:alpha}
\Cref{fig:app-alpha} varies $\alpha$ with all other hyperparameters fixed at the best configuration of each model and dataset.
The norms are reported for the FFN pairs, which carry the largest $\lVert D\rVert_F$ and $\lVert UDV\rVert_{\text{op}}$ of the network at $\alpha = 10^{-2}$.
The best $\alpha$ lies between $3\times10^{-3}$ and $10^{-2}$ in all four settings, and accuracy drops steadily beyond $10^{-2}$.
When the other hyperparameters are re-selected at $\alpha = 10^{-2}$, accuracy stays within 0.3 points of the best in every setting.
Across this range, $\alpha$ trades two quantities against each other.
The norm $\lVert D\rVert_F$ falls as $1/\alpha$ while $\lVert UDV\rVert_{\text{op}}$ grows almost in proportion to $\alpha$, and the attention pairs follow the same trends with a slower growth of the operator norm.
Below $10^{-2}$, a smaller $\alpha$ thus inflates $D$ by orders of magnitude without a consistent gain in accuracy.
Above $10^{-2}$, a larger $\alpha$ shrinks $D$ only by moving scale into the pinned factors, which enlarges the operator norm and lowers accuracy.
We therefore take $\alpha = 10^{-2}$, the largest value inside the accuracy plateau of every setting.

\begin{figure}[htb!]
  \centering
  \includegraphics{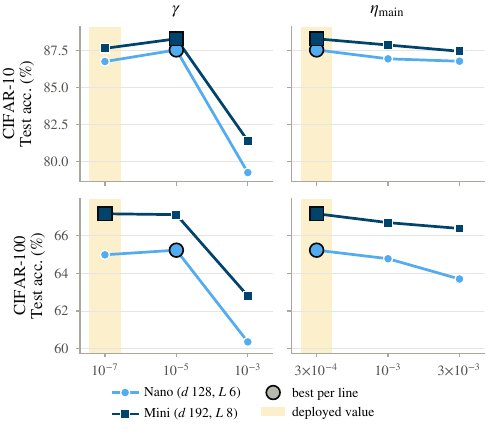}
  \caption{\textbf{Effect of the regularizer strength and the main learning rate.}
  Best test accuracy at each value with all other hyperparameters re-selected.
  Left. Regularizer strength $\gamma$.
  Right. Learning rate $\eta_{\text{main}}$ of the non-UDV parameters.}
  \label{fig:app-gamma}
\end{figure}

\paragraph{Regularizer strength.}
\label{app:regularizer}
In \Cref{fig:app-gamma} (left), the strongest value $\gamma = 10^{-3}$ is clearly harmful, while the two weaker values are close.
At $\gamma = 0$, Nano collapses to 10.4\% on CIFAR-10, and ViT-S on ImageNet falls to chance after 20 epochs.
On ImageNet, $\gamma = 10^{-7}$ also outperforms $10^{-5}$, with 78.01\% against 77.45\%.
We therefore fix $\gamma = 10^{-7}$, the smallest nonzero value of the grid.
For the remaining parameters, $\eta_{\text{main}} = 3\times10^{-4}$ is best in all four settings (\Cref{fig:app-gamma}, right), and we fix it for a batch size of 512.

\begin{figure}[htb!]
  \centering
  \includegraphics[width=\textwidth]{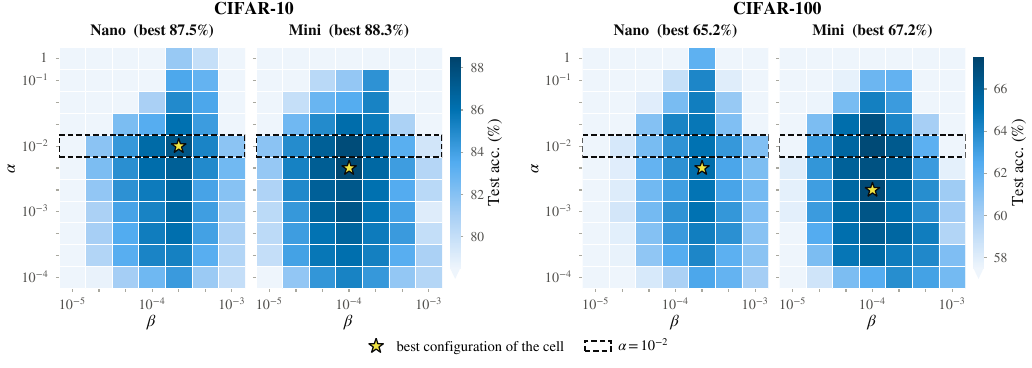}
  \caption{\textbf{Accuracy over the pinning scale and the scaler.}
  Test accuracy of Single-D with per-model $\beta$ on CIFAR, with $\gamma$ and $\eta_{\text{main}}$ re-selected for each pair of $\alpha$ and $\beta$.}
  \label{fig:app-heatmap}
\end{figure}

\begin{figure}[htb!]
  \centering
  \includegraphics{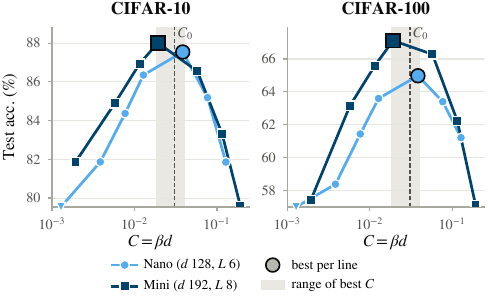}
  \caption{\textbf{Accuracy against the implied constant.}
  Test accuracy of Single-D at $\alpha = 10^{-2}$ against $C = \beta d$, with $\gamma$ and $\eta_{\text{main}}$ re-selected for each $\beta$.}
  \label{fig:app-implied-c}
\end{figure}

\paragraph{Width scaling of the scaler.}
\Cref{fig:app-heatmap} shows the test accuracy over $\alpha$ and $\beta$.
High accuracy forms a band that is wide along $\alpha$ and narrow along $\beta$.
On both datasets the best $\beta$ is $3\times10^{-4}$ for Nano and $10^{-4}$ for Mini, and thus decreases as the model widens.
\Cref{fig:app-implied-c} replots the accuracy at $\alpha = 10^{-2}$ against $C = \beta d$.
In these coordinates the peaks of the two models lie one grid step apart, at $C = 0.0384$ for Nano and $C = 0.0192$ for Mini.
The product $\beta d$ is therefore constant up to the resolution of the grid, which suggests the rule $\beta = C/d$ with a single constant for all models.
As adjacent grid values differ by a factor of 2 to 3, the grid locates $C$ only within this band, and we took $C_0 = 0.0307$ inside it as the initial constant.

\begin{figure}[htb!]
  \centering
  \includegraphics[width=\textwidth]{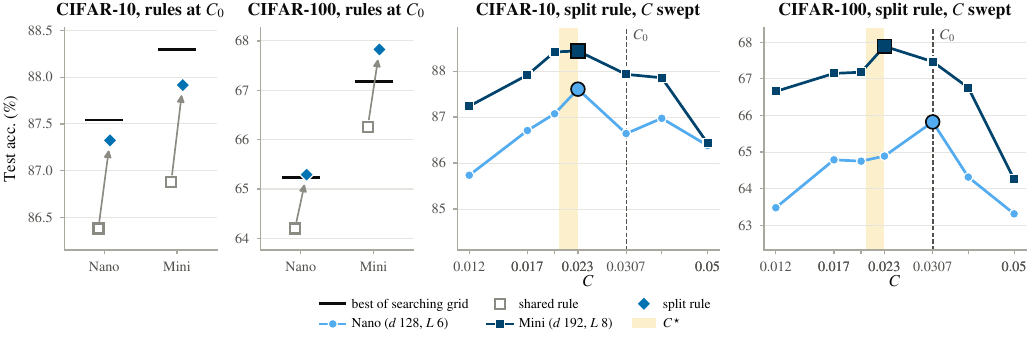}
  \caption{\textbf{Learning-rate rules without search.}
  Single-D at $\alpha = 10^{-2}$ and $\gamma = 10^{-7}$.
  Left two panels. Each arrow goes from the shared rule to the split rule at $C_0 = 0.0307$, and the black bar marks the best of the grid in \Cref{tab:app-grid}.
  Right two panels. Split rule with $C$ swept.
  The dashed line marks $C_0$, and the shaded band marks $C^\star$ from \Cref{app:c-star}.}
  \label{fig:app-rules}
\end{figure}

\paragraph{Splitting the attention rate.}
\Cref{fig:app-rules} (left) first applies the shared rule at $C_0$ with $\alpha = 10^{-2}$ and $\gamma = 10^{-7}$, without any search.
It trails the best configuration of the grid in \Cref{tab:app-grid} by about one point in every setting.
Under the shared rule the attention rate $C\alpha/\sqrt{h}$ falls as the number of heads grows, while the FFN rate stays at $2C\alpha$.
Attention in wider models therefore trains more slowly.
The split rule removes this dependence, improves on the shared rule in all four settings, and reaches the level of the best grid configuration without any search.
\Cref{fig:app-rules} (right) sweeps $C$ under the split rule.
The best $C$ is 0.023 in three of the four settings, which places $C_0$ at the upper end of the good range.
We keep the split rule $\eta_W = C\lVert W\rVert_F/d$ and select $C$ again for the final FFN form over a wider range of widths and depths (\Cref{app:c-star}).

\subsection{FFN forms}
\label{app:ffn-variants}
\begin{figure}[htb!]
  \centering
  \includegraphics[width=\textwidth]{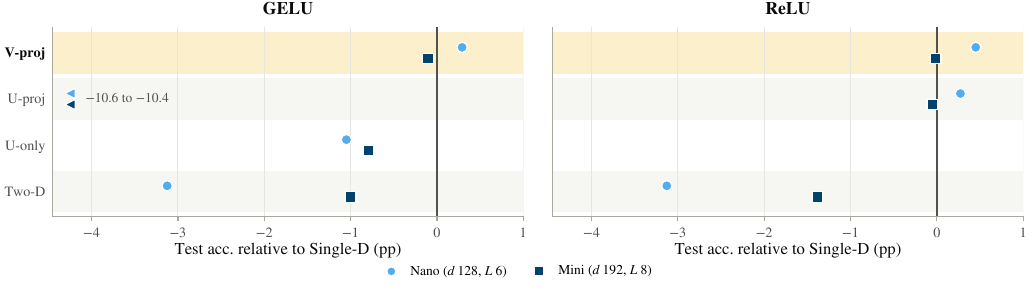}
  \caption{\textbf{Ablation of FFN forms.}
  Test accuracy on CIFAR-10 relative to Single-D.
  The highlighted row is the form used by UDVR.}
  \label{fig:app-ffn-forms}
\end{figure}

\Cref{fig:app-ffn-forms} compares the FFN forms of \Cref{tab:app-ffn-forms} on CIFAR-10 with GELU and ReLU, using Nano and Mini at their best per-model configuration from \Cref{app:old-method}.
Every form starts from the initial function of Single-D and applies its own rebalance from the first step.
With GELU, only V-proj stays close to Single-D, and UDVR therefore uses V-proj.

\subsection{Selecting the constant for UDVR}
\label{app:c-star}
\begin{figure}[htb!]
  \centering
  \includegraphics[width=\textwidth]{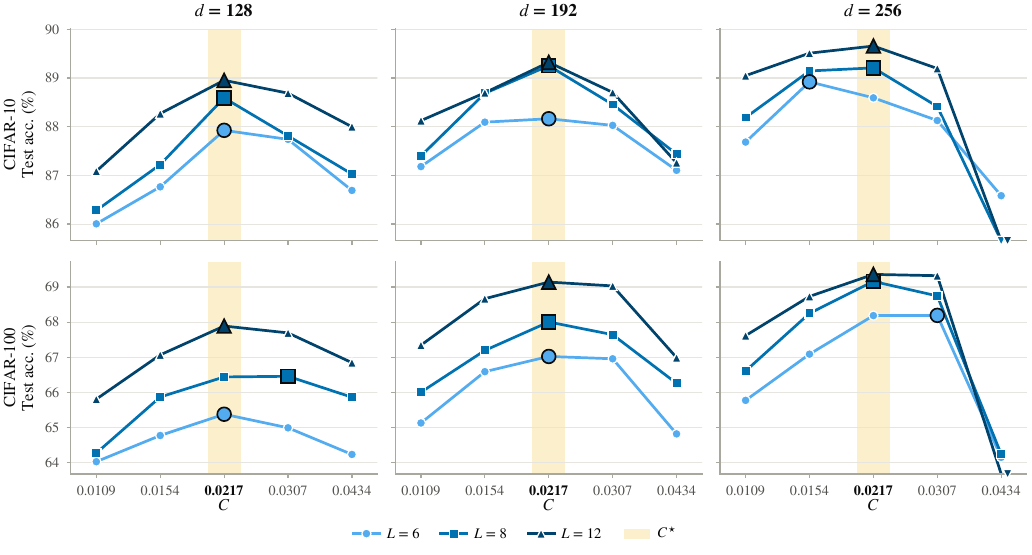}
  \caption{\textbf{Selection of the constant.}
  Test accuracy of UDVR against $C$.
  Columns are widths, rows are datasets, and each line is one depth.
  The shaded band marks $C^\star$, filled markers denote the best $C$ of each line, and triangles mark accuracies below the axis.}
  \label{fig:app-c-selection}
\end{figure}

With the FFN form fixed, we selected $C$ once for UDVR under the split rule, keeping $\alpha$, $\gamma$, and $\eta_{\text{main}}$ from \Cref{app:old-method}.
The sweep in \Cref{fig:app-c-selection} covers all CIFAR ViTs of \Cref{tab:models} on both datasets, giving 18 settings that include Nano and Mini.
Drop path is 0.1 for $L = 12$ and zero otherwise.
The constant takes five values $C = C_0\,2^{k/2}$ for $k = -3, \dots, 1$.

In almost every setting the best value is $C^\star = C_0/\sqrt{2} = 0.0217$, an interior point of the grid, with no systematic dependence on width or depth.
We therefore use $C^\star$ as the single constant of UDVR.

\section{Additional experiments and results}
\label{app:addresult}

\subsection{Wider and deeper models}
\label{app:cifar_widedeep}

\begin{figure}[htb!]
  \centering
  \includegraphics[width=\textwidth]{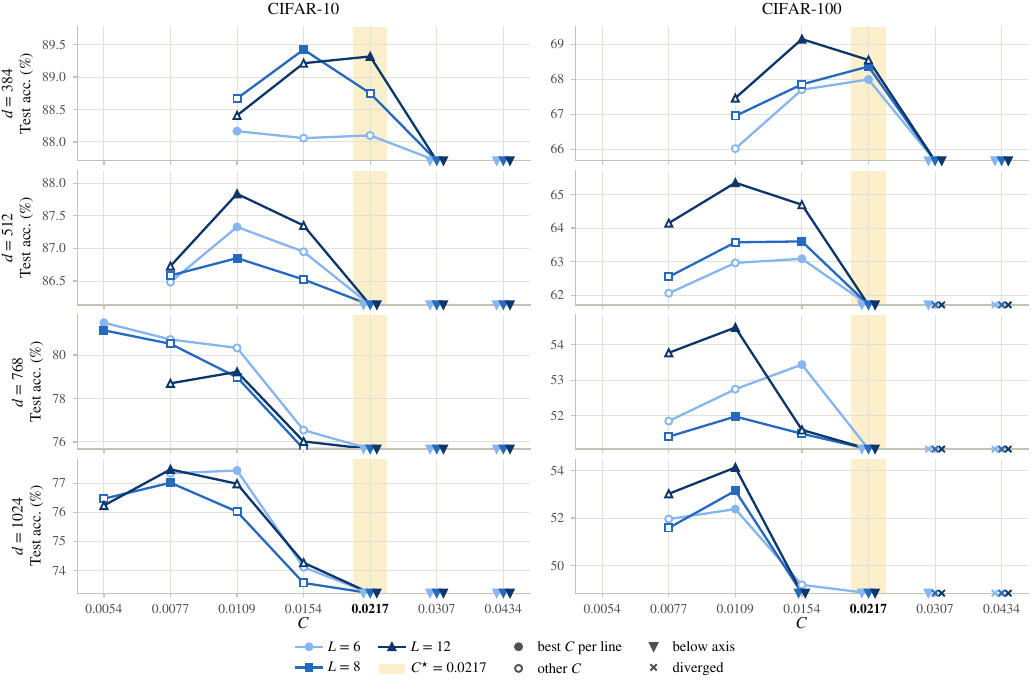}
  \caption{\textbf{The learning rate scaler constant on wider models.}
  Test accuracy of UDVR against $C$ on CIFAR with one panel per width and dataset and one line per depth.}
  \label{fig:app-wide-c}
\end{figure}

\para{Wider models on CIFAR.}
We extend the CIFAR sweep of $C$ to wider models with $d \in \{384, 512, \allowbreak 768, 1024\}$ at depths $L \in \{6, 8, 12\}$, where the training recipe is unchanged.
\Cref{fig:app-wide-c} shows that $C^\star$ stays close to the best $C$ of every model up to $d = 384$, but from $d = 512$ on it incurs a large accuracy loss as the optimum shifts to smaller values.
Accuracy decays only slowly when $C$ is too small but collapses past a cliff on the large-$C$ side, and this cliff moves toward smaller $C$ as the model widens until it reaches $C^\star$ itself at $d \ge 512$.
Halving $C$ as a diagnostic brings every wide model back to near its best accuracy, which indicates that the failure at large width is instability rather than slow training.

\begin{figure}[htb!]
  \centering
  \includegraphics[width=\textwidth]{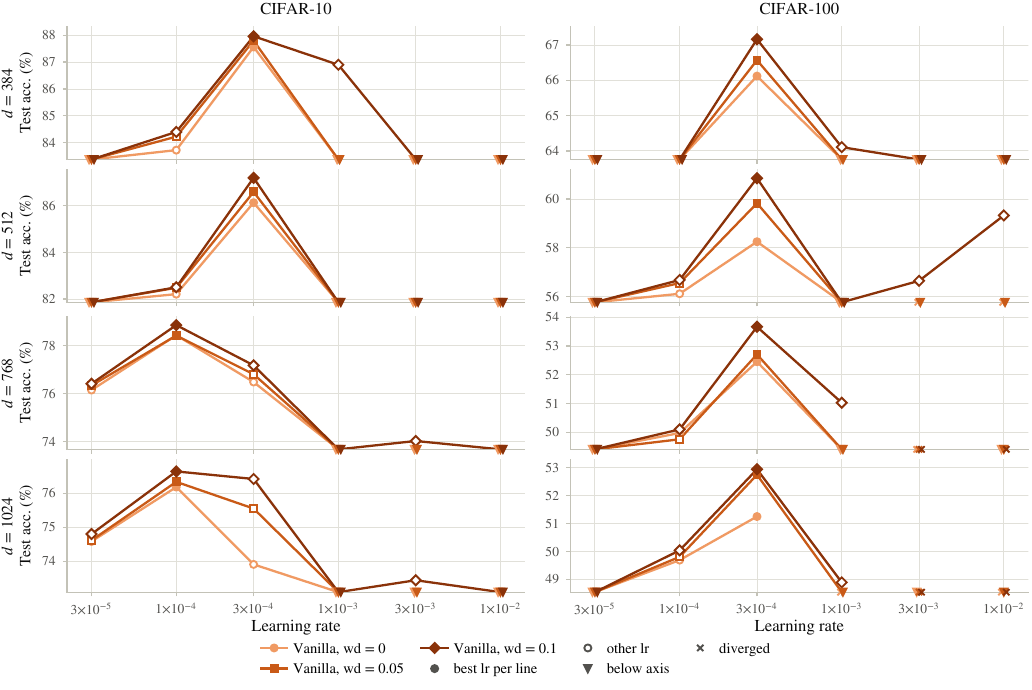}
  \caption{\textbf{Tuned Vanilla on wider models.}
  Test accuracy of Vanilla at depth 12 against the learning rate and weight decay on CIFAR.}
  \label{fig:app-wide-vanilla}
\end{figure}

\begin{figure}[htb!]
  \centering
  \includegraphics{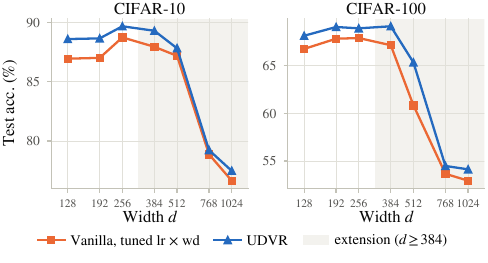}
  \caption{\textbf{Accuracy comparison against width.}
  Test accuracy of UDVR at its best $C$ and tuned Vanilla on CIFAR at depth 12.}
  \label{fig:app-wide-trend}
\end{figure}

However, this accuracy loss with width is not specific to UDVR.
\Cref{fig:app-wide-trend} shows that accuracy falls beyond $d = 384$ even at each model's best $C$, and Vanilla tuned at every width loses a comparable amount.
Vanilla's best learning rate also shrinks with width and its high learning rates collapse on the wide models (\Cref{fig:app-wide-vanilla}), mirroring the cliff of UDVR.
With $C$ re-tuned, UDVR stays above tuned Vanilla at every width.

We consider three possible causes for this degradation:
\begin{redlist}
\item \textbf{Too little data for the model size.}
The parameter count grows quadratically with $d$ while the training set stays at 50k images, and ViTs have weak inductive biases.
Training accuracy drops far less than test accuracy on both datasets, indicating that most of the loss comes from a widening generalization gap.
\item \textbf{Regularization tuned for narrow models.}
The recipe keeps augmentation and regularization fixed across widths, and tuned Vanilla picks the largest available weight decay at every $d \ge 384$, suggesting that wide models need stronger regularization.
\item \textbf{A narrower range of stable step sizes.}
When updates are coherent across the entries of a matrix, as Adam updates typically are, a fixed relative step $C$ changes the layer output roughly as $C\sqrt{d}$ and thus becomes more aggressive with width.
This matches both the leftward shift of the cliff and the shrinking best learning rate of Vanilla.
\end{redlist}
The last cause alone would predict the same collapse on any dataset, yet with the same $C^\star$ ViT-B trains stably on ImageNet and outperforms ViT-S (\Cref{fig:baseline-imagenet}), whereas the same shape loses heavily on CIFAR even at its best $C$.
With about $25\times$ more training images on ImageNet, the stable range appears to depend on data volume, pointing to data volume and regularization as the main factors.
However, this comparison does not isolate data volume since ImageNet also differs in resolution, augmentation and batch size.

\para{Deeper models on CIFAR.}
The rule for $C$ has no depth factor, and the sweep that selected $C^\star$ covered only $L \le 12$.
We therefore keep $C^\star$ fixed and increase the depth up to $L = 32$ at widths $d \le 256$, where $C^\star$ is not close to the stability edge.
All other settings follow the CIFAR recipe of the main text except that every depth uses drop-path $0.1$.

\begin{figure}[htb!]
  \centering
  \includegraphics{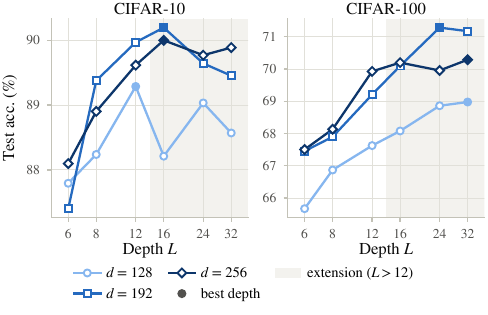}
  \caption{\textbf{Test accuracy against depth.}
  Test accuracy of UDVR at the fixed $C^\star$ on CIFAR.}
  \label{fig:app-depth-acc}
\end{figure}

\begin{figure}[htb!]
\centering
\includegraphics{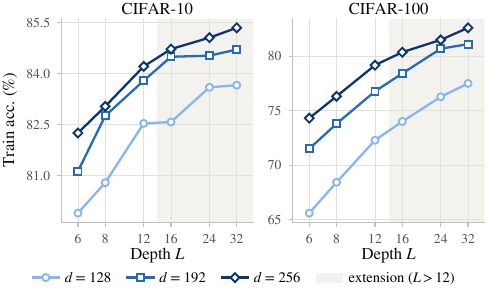}
\caption{Training accuracy of UDVR at the fixed $C^\star$ on CIFAR.
Training accuracy is measured in training mode with augmentation, mixup and cutmix applied.}
\label{fig:app-depth-train-acc}
\end{figure}

\Cref{fig:app-depth-acc} shows that depth helps every model up to $L = 12$ on both datasets.
Beyond $L = 12$, accuracy keeps rising on CIFAR-100 but saturates on CIFAR-10.
This saturation is not an optimization limit.
Training accuracy rises monotonically with depth in every model (\Cref{fig:app-depth-train-acc}), but only on CIFAR-100 does test accuracy follow (\Cref{fig:app-depth-acc}).
With the same number of training images in both datasets, this suggests that the gain from extra capacity is bounded by the data, as observed for width.

\subsection{UDVR with weight decay on non-UDV parameters}
\label{app:udvr_wd}

UDVR trains its reparameterized pairs without weight decay, and by default the remaining parameters (the main group) use none either.
We test adding weight decay 0.05 to the main group while keeping the UDVR pairs at zero, with each run differing from the rule-point run of the same model only in this setting.
\Cref{tab:app-mainwd} shows that the sign of the effect differs between the two sizes of ViT and of Swin.
A nonzero value would thus be a model-specific choice outside the tuning-free setting, and we keep zero weight decay on the main group as the default.

\begin{table}[th]
\centering
\small
\caption{ImageNet test accuracy (\%) of UDVR with weight decay 0 and 0.05 on the main group, which contains all non-UDV parameters (embeddings, normalization layers, biases, relative position biases and the classifier head, plus the MBConv blocks and the stem in MaxViT).
Main (\%) is the share of parameters in the main group.}
\begin{tabular}{cc c ccc}
\toprule
Model & Size & Main (\%) & wd 0 & wd 0.05 & $\Delta$ \\
\midrule
\multirow{2}{*}{ViT}    & S & 3.5  & \textbf{78.63} & 78.57 & $-0.06$ \\
                        & B & 1.8  & 79.97 & \textbf{80.23} & $+0.27$ \\
\midrule
\multirow{2}{*}{Swin}   & T & 8.4  & 79.37 & \textbf{79.67} & $+0.30$ \\
                        & B & 4.5  & \textbf{82.16} & 81.46 & $-0.70$ \\
\midrule
\multirow{2}{*}{MaxViT} & T & 30.0 & 82.37 & \textbf{82.77} & $+0.39$ \\
                        & S & 29.4 & 82.98 & \textbf{83.53} & $+0.55$ \\
\bottomrule
\end{tabular}%
\label{tab:app-mainwd}
\end{table}

\subsection{SwiGLU activation}
\label{app:swiglu}

We replace the GELU feed-forward network with SwiGLU, $y = (\mathrm{SiLU}(x W_{\mathrm{gate}}) \odot x W_{\mathrm{up}})\, W_{\mathrm{down}}$, with hidden width $\tfrac{2}{3} d_{\mathrm{mlp}}$ to match the parameter count and FLOPs of the GELU model.
In UDVR, the GELU feed-forward network is the pair $x \mapsto \mathrm{GELU}(x U D)\, V$.
With SwiGLU we keep this pair for the up and down projections and add the gate as a second $U$ with its own diagonal, $y = \big(\mathrm{SiLU}(x U_{g} D_{g}) \odot x U D\big) V$.
$U_g$ is treated like $U$, with its columns pinned and their scale moved into $D_g$, which is regularized like $D$.
The learning rates follow from the rule for the new shapes, and Vanilla keeps the learning rate and weight decay of its GELU optimum, leaving neither method retuned for SwiGLU.

\Cref{tab:app-swiglu} shows that SwiGLU improves both methods on both models, with UDVR gaining about twice as much as Vanilla.
This narrows the gap to Vanilla, although UDVR stays below it.
The extra gain of UDVR is largest early in training and shrinks as the learning rate decays.
The deployment rule thus carries over to a gated feed-forward network without changes.

\begin{table}[th]
\centering
\small
\caption{ImageNet test accuracy (\%) with GELU and SwiGLU feed-forward networks.}
\begin{tabular}{c ccc ccc}
\toprule
 & \multicolumn{3}{c}{Vanilla} & \multicolumn{3}{c}{UDVR} \\
\cmidrule(lr){2-4}\cmidrule(lr){5-7}
Model & GELU & SwiGLU & $\Delta$ & GELU & SwiGLU & $\Delta$ \\
\midrule
ViT-S  & 79.82 & \textbf{80.12} & $+0.30$ & 78.63 & 79.26 & $+0.63$ \\
Swin-T & 80.41 & \textbf{80.84} & $+0.43$ & 79.37 & 80.21 & $+0.83$ \\
\bottomrule
\end{tabular}%
\label{tab:app-swiglu}
\end{table}

\subsection{Comparison with muP on hyperparameter transfer}
\label{app:mup}

$\mu$P \cite{yang2021tuning} offers Vanilla its own way to transfer a learning rate across widths without tuning.
The original method tunes the proxy on the same data as the target model, and its authors note that the transferred hyperparameters may not hold across datasets.
Our cross-dataset protocol, which UDVR follows as well, is thus an extended use of $\mu$P.
We select $\eta^\star = 2.8\times10^{-3}$ on a CIFAR proxy of width 128 and depth 12 by the criterion used for $C^\star$ and deploy it on ImageNet without search or batch-size rescaling.
With width multiplier $m = d/128$ (per stage for Swin), hidden matrices use learning rate $\eta/m$, weight decay $0.05\,m$ and initialization scaled by $1/\sqrt{m}$ relative to the proxy, vector-like parameters use $\eta$ and weight decay 0.05, and the readout input is scaled by $1/m$.
Scaling the weight decay with $m$ keeps the per-step decay $\eta\lambda$ at $0.05\,\eta$ for every width.

\Cref{tab:app-mup} shows that the two schemes are on par on ViT-S, ViT-B and Swin-T, whose depth stays close to that of the proxy.
On Swin-B, whose third stage has 18 blocks, $\mu$P falls well below both UDVR and tuned Vanilla, while UDVR stays close to tuned Vanilla.
The $\mu$P model underfits, with lower training accuracy than UDVR and Vanilla and a test accuracy that stalls in the middle of training before recovering as the learning rate decays.
This suggests a step size too large for the deep stage.
$\mu$P thus transfers well across width at the depth of the proxy but not to a substantially deeper model, consistent with its theory covering width only.
The rule for $C$ has no depth correction either, yet UDVR tolerates the deeper model, consistent with the depth study (\Cref{app:cifar_widedeep}).
Within both model families, the difference moves in favor of UDVR from the smaller to the larger model, suggesting that the advantage of UDVR grows with model size.

\begin{table}[htb!]
\centering
\small
\caption{ImageNet test accuracy (\%) of width-transfer schemes without ImageNet search.}
\begin{tabular}{c cc c c}
\toprule
 & \multicolumn{2}{c}{No ImageNet search} & Grid search & \\
\cmidrule(lr){2-3}\cmidrule(lr){4-4}
Model & $\mu$P & UDVR & Vanilla & $\mu$P $-$ UDVR \\
\midrule
ViT-S  & \textbf{79.22} & 78.63 & 79.82 & $+0.59$ \\
ViT-B  & 79.84 & \textbf{79.97} & 80.12 & $-0.13$ \\
Swin-T & \textbf{79.39} & 79.37 & 80.41 & $+0.02$ \\
Swin-B & 79.90 & \textbf{82.16} & 82.49 & $-2.26$ \\
\bottomrule
\end{tabular}%
\label{tab:app-mup}
\end{table}

\subsection{Transfer of the predefined constant to other LLM architectures}
\label{app:transferC_morellm}

To extend the language-model results beyond the GPT-style architecture of the main text, we train two further architectures at the small scale (\Cref{app:tab llm_other_models,app:tab llm_hyperparams_more}).
Llama-S replaces LayerNorm and the GELU feed-forward network with RMSNorm and SwiGLU, and OLMoE-S further replaces the feed-forward network with a mixture of 64 SwiGLU experts with eight active per token, matching the active hidden width of Llama-S.
Both omit QK-norm and the reordered normalization of OLMoE, which would normalize away the scale that UDVR moves into $D$.
Data, token budget and training protocol follow the main text.

UDVR uses the rule point of the main text with $C^\star$ taken from the vision experiments and is trained once per model.
Each SwiGLU block, including every expert of OLMoE-S, is reparameterized as in \Cref{app:swiglu} with its own diagonals, while the router stays in the main group.
The feed-forward learning rate follows from the rule with the active hidden width.
On Llama-S, Vanilla, WN and $\sigma$Reparam are tuned over the learning rate, and Vanilla also over the weight decay.
On OLMoE-S, each baseline uses a single point, the optimum shared by the three model sizes of the main text.

\Cref{tab:app-morellm} shows that on Llama-S UDVR is close to tuned Vanilla in validation perplexity, much closer than on the GPT-style model of the same size, and outperforms tuned WN and $\sigma$Reparam.
On OLMoE-S, UDVR has the lowest perplexity, but the baselines there are not tuned and we do not claim that UDVR beats them.
The gain from the mixture of experts is similar for all methods, and routing stays as balanced under UDVR as under the baselines.
On LAMBADA, UDVR trails Vanilla slightly and is level with WN and $\sigma$Reparam, the same pattern as for the GPT-style model.
Overall, the deployment rule transfers without changes to an RMSNorm and SwiGLU architecture and to a mixture of experts.

\begin{table}[h]
\centering
\caption{Llama- and OLMoE-style model configurations.}
\resizebox{\textwidth}{!}{%
\begin{tabular}{lcccccccr}
\toprule
Model & Depth & Heads & $d$ & $d_{\text{head}}$ & Norm & FFN & $d_{\text{mlp}}$ & Trainable \\
\midrule
Llama-S & 22 & 8 & 512 & 64 & RMSNorm & SwiGLU                 & 1536            & 126.51M \\
OLMoE-S & 22 & 8 & 512 & 64 & RMSNorm & SwiGLU MoE (top-8)     & $64 \times 192$ & 490.56M (127.23M active) \\
\bottomrule
\end{tabular}%
}
\label{app:tab llm_other_models}
\end{table}

\begin{table}[htb]
\centering
\caption{Pre-training hyperparameters and configuration for the Llama-style and OLMoE-style models.}
\renewcommand{\arraystretch}{1.15}
\resizebox{\textwidth}{!}{%
\begin{tabular}{@{}llll@{}}
\toprule
\textbf{Category} & \textbf{Item} & \textbf{Llama-S} & \textbf{OLMoE-S} \\
\midrule
\multirow{4}{*}{Optimizer}
 & Optimizer & \multicolumn{2}{l}{AdamW} \\
 & Betas & \multicolumn{2}{l}{(0.9, 0.95)} \\
 & Eps & \multicolumn{2}{l}{1e-8} \\
 & Gradient clip & \multicolumn{2}{l}{None} \\
\midrule
\multirow{2}{*}{LR schedule}
 & Scheduler & \multicolumn{2}{l}{linear warmup then cosine decay to 0, stepped per optimizer step} \\
 & Warmup ratio & \multicolumn{2}{l}{0.01} \\
\midrule
\multirow{8}{*}{Data}
 & Corpus & \multicolumn{2}{l}{SlimPajama-627B} \\
 & Tokenizer & \multicolumn{2}{l}{GPT-NeoX BPE} \\
 & Domains & \multicolumn{2}{l}{arxiv, book, c4, commoncrawl, github, stackexchange, wikipedia} \\
 & Sequence length & \multicolumn{2}{l}{2048} \\
 & Vocabulary size & \multicolumn{2}{l}{50,280 (padded to 50,304)} \\
 & Global batch size & \multicolumn{2}{l}{128 sequences} \\
 & (GPU, micro-batch, grad-accum) & (1, 8, 16) & (4, 4, 8) \\
 & Token budget & \multicolumn{2}{l}{7e9 (uniform 1e9 $\times$ 7 domains, single pass)} \\
\midrule
Seed & Random seed & \multicolumn{2}{l}{42 (the other two random seeds are derived from this seed)} \\
\midrule
Precision & Numerical precision & \multicolumn{2}{l}{Autocast bfloat16} \\
\midrule
\multirow{5}{*}{Architecture}
 & Positional encoding & \multicolumn{2}{l}{RoPE (base 10,000)} \\
 & Attention & \multicolumn{2}{l}{Multi-head, no QK-norm} \\
 & Normalization & \multicolumn{2}{l}{RMSNorm (eps 1e-5, weight only), pre-norm} \\
 & FFN & SwiGLU, hidden 1536 & 64 SwiGLU experts, hidden 192 each, top-8, dropless (grouped GEMM) \\
 & Embedding & \multicolumn{2}{l}{Untied} \\
\midrule
\multirow{2}{*}{Router}
 & Router & -- & Linear $64 \times 512$, no bias, logits in fp32 \\
 & Expert weights & -- & Softmax over all 64 experts, top-8 probabilities used without renormalization \\
\midrule
\multirow{2}{*}{Objective}
 & Train criterion & Next-token cross-entropy & Next-token cross-entropy $+$ 0.01 load-balance loss $+$ 0.001 router z-loss \\
 & Eval criterion & \multicolumn{2}{l}{Next-token cross-entropy only, aggregated per domain} \\
\midrule
Evaluation & Eval subset size & \multicolumn{2}{l}{1.5e7 tokens per domain, unused in training} \\
\bottomrule
\end{tabular}%
}
\label{app:tab llm_hyperparams_more}
\end{table}

\begin{table}[htb!]
\centering
\small
\caption{Validation perplexity and LAMBADA accuracy of Llama-S and OLMoE-S.}
\begin{tabular}{l cc cc}
\toprule
 & \multicolumn{2}{c}{Val.\ PPL $\downarrow$} & \multicolumn{2}{c}{LAMBADA (\%) $\uparrow$} \\
\cmidrule(lr){2-3}\cmidrule(lr){4-5}
Method & Llama-S & OLMoE-S & Llama-S & OLMoE-S \\
\midrule
UDVR            & 12.432          & \textbf{10.897} & 24.98          & 28.57 \\
Vanilla         & \textbf{12.404} & 10.939          & \textbf{25.91} & 29.32 \\
WN              & 12.617          & 11.202          & 24.88          & 28.24 \\
$\sigma$Reparam & 12.674          & 11.258          & 24.84          & \textbf{29.36} \\
\bottomrule
\end{tabular}%
\label{tab:app-morellm}
\end{table}

\end{document}